\documentclass[letterpaper]{article} 
\usepackage[preprint]{aaai2027} 
\usepackage[hyphens]{url} 
\usepackage{graphicx} 
\usepackage{natbib} 
\usepackage{caption} 
\usepackage{amsmath,amssymb}
\usepackage{booktabs}
\usepackage{array}
\usepackage{multirow}
\usepackage{algorithm}
\usepackage{algpseudocode}
\usepackage[capitalise]{cleveref}
\usepackage[table]{xcolor}
\usepackage{xcolor}
\colorlet{MyPink}{orange!8}
\colorlet{MyCream}{yellow!8}

\definecolor{paradeyellow}{RGB}{255,249,219}

\newcommand{\std}[1]{%
  {\fontsize{6.7pt}{6.5pt}\selectfont$\pm$#1}%
}

\newcommand{\method}{PARADE}
\newcommand{\tasklora}{Task LoRA}
\newcommand{\demolora}{Demographic LoRA}

\newcolumntype{L}[1]{>{\raggedright\arraybackslash}p{#1}}
\newcolumntype{C}[1]{>{\centering\arraybackslash}p{#1}}
\newcommand{\selectedcell}[1]{\cellcolor{paradeyellow}#1}

\title{Composable Parametric Demographic Injection for Values Survey Simulation}
\author{
Xinyuan Cao\textsuperscript{\rm 2,\rm 1},
Qingyi Pan\textsuperscript{\rm 2},
Bangde Du\textsuperscript{\rm 2},
Yujia Zhou\textsuperscript{\rm 2},
Hexi Wang\textsuperscript{\rm 2},\\
Weihang Su\textsuperscript{\rm 2},
Yiqun Liu\textsuperscript{\rm 2},
Yueyue Wu\textsuperscript{\rm 2},
Min Zhang\textsuperscript{\rm 2},
Qingyao Ai\textsuperscript{\rm 1,\rm 2}\corresponding
}

\affiliations{
\textsuperscript{\rm 1}Quancheng Laboratory\\
\textsuperscript{\rm 2}Department of Computer Science and Technology, Tsinghua University, Beijing 100084, China\\
cao-xy22@mails.tsinghua.edu.cn, aiqy@tsinghua.edu.cn
}

\begin{document}

\maketitle

\begin{abstract}

Large language models (LLMs) are increasingly used as simulated respondents in computational social science, yet accurately reproducing population-specific heterogeneous response distributions remains challenging. Existing methods such as demographic prompting are sensitive to prompt design and increase inference overhead, while parameter-based methods struggle to reuse demographic information efficiently,
limiting generalization to unseen combinations.
In this regard, we propose \textbf{\method{}}, a composable \textbf{PARA}metric \textbf{DE}mographic injection framework that represents each demographic attribute value with a reusable and composable LoRA module. 
Specifically, \method{} separates task-common survey capability from demographic-specific value shifts via a shared Task LoRA and attribute-level Demographic LoRAs, with composition calibration to reduce parameter interference. 
At inference time, target population models can be constructed on demand by recombining the relevant modules, including demographic combinations not observed during training.
Extensive experiments on the World Values Survey and SocioBench across three backbone models show that \method{} achieves strong population-level distribution alignment while maintaining competitive predictive performance, reducing KL divergence by up to 52.8\% relative to the strongest baseline. Code is available at https://github.com/zerzer-studying/PARADE.git.
\end{abstract}

\section{Introduction}

Large language models (LLMs) are increasingly used as simulated respondents in computational social science \citep{achiam2023gpt,argyle2023out,aher2022using,kim2023ai,ashokkumar2026large}. Such simulations can support survey pretesting, large-scale behavioral analysis, and the exploration of social hypotheses before costly human studies.
The value of such simulations depends on whether they reproduce empirical response distributions for clearly defined target populations. 
Because human values vary systematically across demographics such as gender, age, and country, a practical value simulator should accurately reproduce the response distributions of specific populations under demographic conditions.

\begin{figure}[t]
    \centering
    \includegraphics[width=\columnwidth]{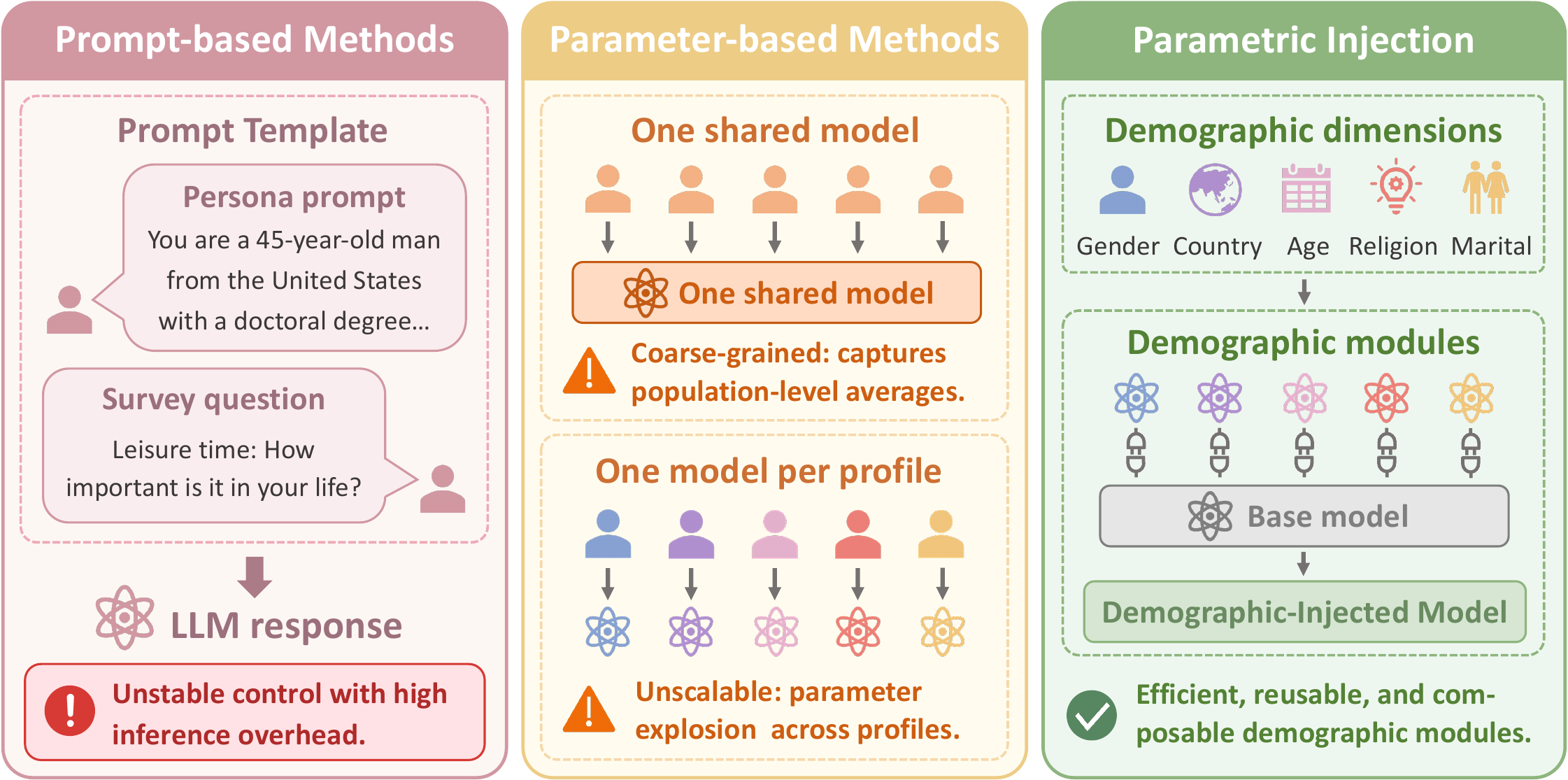}
    \caption{
        Comparison between existing methods and our composable parametric demographic injection framework.
    }
    \label{fig:motivation}
\end{figure}

As illustrated in~\cref{fig:motivation}, existing studies primarily rely on persona prompting or parameter adaptation to condition LLMs on target populations. Persona prompting requires no additional training, but the target profile must be repeatedly included for every survey question, increasing inference overhead and making model behavior sensitive to prompt wording, formatting, and demographic stereotypes inherited from pretraining \citep{lutz2025prompt,hu2024quantifying,sun2025sociodemographic,cheng2023marked}. 
On the other hand, parameter-based methods can learn group-specific differences directly from observed survey responses \citep{huang2026distribution,sun2026cuma,wang2026parametric}, but they still face an inherent trade-off between modeling fidelity and scalability. Pooling all groups into a single model is efficient, yet may average out meaningful value differences across populations \citep{cao2025specializing,suh2025language}.
Training a separate adapter for every complete demographic profile can better preserve specific value patterns, but the number of parameters grows rapidly with the number of demographic combinations \citep{tan2024democratizing}, and such methods cannot support profiles absent during training.
Based on this analysis, the main challenge is to represent multidimensional demographic profiles in a reusable and compositional manner while preserving group-specific value patterns.

To address this challenge, we propose \textbf{\method{}}, a composable \textbf{PARA}metric \textbf{DE}mographic injection framework that factorizes each multidimensional demographic profile into reusable low-rank adaptation (LoRA) modules associated with individual attribute values.  
The key idea is that demographic profiles are composed of multiple attribute values, whose effects can be shared across populations with the same attributes.
Therefore, \method{} represents demographic effects at the attribute level rather than learning separate parameters for complete profiles.
Specifically, \method{} first trains a shared \textit{Task LoRA} to capture the shared survey-response protocol, thereby avoiding redundant encoding of shared protocol information.
It then jointly trains activated \textit{Demographic LoRAs} to capture specific patterns of demographic value.
We further use a soft orthogonality regularizer that reduces interference between task and demographic modules, and learnable dimension-level weights to calibrate contributions during demographic composition.
At inference time, \method{} combines the attribute modules corresponding to the target profile with the shared task module. 
Therefore, \method{} can construct new population models by recombining observed demographic attributes, including complete demographic combinations not encountered during training.

To evaluate the effectiveness of \method{}, we conduct extensive experiments on the World Values Survey and SocioBench across three backbone models. The empirical results show that \method{} achieves stronger alignment with population-level response distributions while maintaining competitive predictive performance, demonstrating the effectiveness of composable attribute-level demographic modeling. Across the three backbones, \method{} reduces KL divergence relative to the strongest baseline by up to
52.8\% on WVS and reduces JS divergence by up to 15.9\% on SocioBench.
Our main contributions are summarized as follows: 
\begin{itemize}
    \item We propose \method{}, a composable parametric injection framework capturing multidimensional demographic profiles through reusable attribute-level modules.
    \item We introduce a two-stage training procedure to separate task-common survey capability from demographic-specific value patterns,  with composition calibration to improve compatibility among demographic modules.
    \item We evaluate \method{} on two survey datasets and three backbones, demonstrating strong distributional fidelity and generalization to unseen demographic combinations.
\end{itemize}

\section{Problem Formulation}
\label{sec:preliminaries}

\subsection{Population Distribution Simulation}
\label{sec:problem_formulation}

We study demographic-conditioned value simulation at the population level, where language models are used to approximate the response distribution of target populations
under different demographic conditions.
Formally, we consider a model parameterized by $\Theta$ and a value survey question $q$ with candidate answer set $\mathcal{A}_q$. Given a target demographic profile $g=\{(d_i,v_i)\}_{i=1}^{m}$, where each $(d_i,v_i)$ specifies a demographic dimension and its value, our objective is to estimate the population-level response distribution \begin{equation} 
p_{\Theta}(a\mid q,g), \qquad a\in\mathcal{A}_q. 
\end{equation}
Since demographic profiles are composed of multiple attribute values, we represent demographic effects through reusable attribute-level components. Such components can be shared across populations and composed to construct different demographic conditions.

\subsection{Composable Demographic Injection}

Based on the attribute-level formulation,
we associate each demographic attribute value with a reusable parameter module rather than assigning independent parameters to complete profiles.
Formally, for an adapted layer $\ell$ with frozen weights
$W_\ell\in\mathbb{R}^{h_{\mathrm{out}}\times h_{\mathrm{in}}}$,
a LoRA module $z$ consists of a down-projection
$A_{\ell,z}\in\mathbb{R}^{r\times h_{\mathrm{in}}}$ and an up-projection
$B_{\ell,z}\in\mathbb{R}^{h_{\mathrm{out}}\times r}$. It induces the
low-rank weight update
\begin{equation}
\Delta W_{\ell,z}
=
\frac{\alpha}{r}B_{\ell,z}A_{\ell,z},
\label{eq:lora_update}
\end{equation}
where $r$ is the LoRA rank and $\alpha$ is the scaling factor. The index $z$ denotes either a demographic attribute value $(d,v)$ or a shared adaptation module $t$.

For a profile containing multiple demographic attributes
$g=\{(d_1,v_1),\ldots,(d_m,v_m)\}$, its composed demographic update at
layer $\ell$ is
\begin{equation}
\Delta W_\ell(g)
=
\sum_{(d,v)\in g}
w_d\Delta W_{\ell,d,v},
\label{eq:demographic_composition}
\end{equation}
where $w_d$ is a learnable weight shared by all attribute values within
demographic dimension $d$. The effective weight used for profile $g$ is obtained by combining the frozen backbone, the shared adaptation module, and the composed demographic modules as
\begin{equation}
\widetilde{W}_\ell(g)
=
W_\ell
+
\Delta W_{\ell,t}
+
\Delta W_\ell(g).
\label{eq:effective_weight}
\end{equation}

This formulation enables the same parameter module to reuse supervision from different profiles containing the corresponding attribute value during training,
and learned modules can be recombined to construct complete profiles not observed during training.

\begin{figure*}[t]
    \centering
\includegraphics[
  width=\linewidth,
  trim={0 37pt 30pt 0},
  clip
]{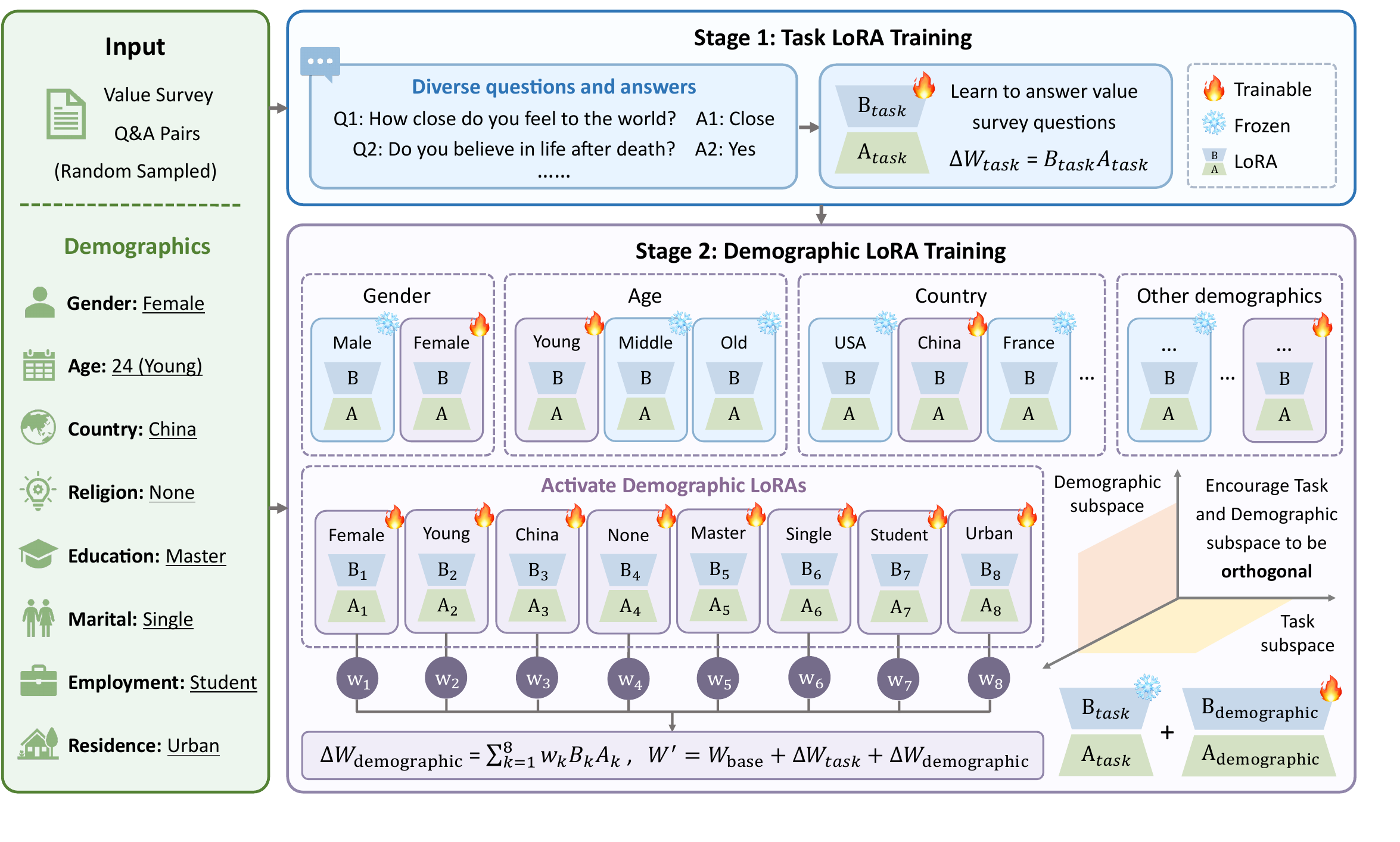}
    \caption{
        Overview of \method{} training. In Stage~1, a shared \tasklora{} is
        trained on demographic-free survey data and then frozen. In
        Stage~2, each demographic attribute value is assigned a reusable
        \demolora{}. Given a respondent profile, only the corresponding
        demographic modules are activated and jointly trained.
        Dimension-level weights calibrate their contributions, while a
        soft orthogonality objective separates demographic adaptation
        from task-common adaptation.
    }
    \label{fig:pipeline}
\end{figure*}

\subsection{Rationale for Additive Composition}
\label{sec:additive_basis}

Because a demographic profile consists of multiple attribute values, we analyze whether their contributions can be approximately decomposed and aggregated to recover the response distribution of the complete profile.

Let $p_0(a\mid q)$ denote the demographic-agnostic response distribution
of the pretrained backbone, and 
$p_{d,v}(a\mid q)=p(a\mid q,d=v)$ denote the response distribution
conditioned on a single demographic attribute value. 
Under the assumption that demographic attributes are conditionally independent given the survey question and response,
we can derive
\begin{equation}
p(a\mid q,g)
\propto
\frac{
\prod_{(d,v)\in g} p_{d,v}(a\mid q)
}{
p_0(a\mid q)^{|g|-1}
}.
\label{eq:probability_composition}
\end{equation}

Let $\mathbf{s}_0(q)$, $\mathbf{s}_{d,v}(q)$, and $\mathbf{s}_g(q)$
denote the corresponding pre-softmax logit vectors, and define the
single-attribute logit shift as
$\Delta\mathbf{s}_{d,v}(q)
=
\mathbf{s}_{d,v}(q)-\mathbf{s}_0(q)$.
Then we can derive
\begin{equation}
\mathbf{s}_g(q)-\mathbf{s}_0(q)
\doteq
\sum_{(d,v)\in g}
\Delta\mathbf{s}_{d,v}(q),
\label{eq:logit_composition}
\end{equation}
where $\doteq$ denotes equality up to an answer-independent additive
constant, which does not affect the softmax distribution. 
Each attribute-specific shift is represented by a corresponding LoRA parameter update. 
Thus, under the conditional-independence assumption, the demographic effect of a complete profile can be approximated by summing its single-attribute logit shifts, providing the rationale for additive demographic module composition. 
This assumption is introduced only to motivate additive composition; our model does not require demographic attributes to be strictly independent in practice.
Detailed proofs are provided in Appendix A.

\section{Our Method}

We propose \method{}, a composable parametric demographic injection framework for value simulation, as shown in~\cref{fig:pipeline}.
\cref{sec:stage1} introduces \tasklora{} to capture task-common survey capability, thereby avoiding its redundant encoding across demographic modules;
\cref{sec:stage2} jointly trains co-activated \demolora{} modules to improve compatibility during composition; 
\cref{sec:combine} describes how the modules can be composed during inference to generate population-level response distributions.

\subsection{Stage 1: Task LoRA Training}
\label{sec:stage1}

Motivated by the additive formulation in
\cref{eq:probability_composition},
the shared survey-response component should be introduced only once rather than repeatedly through every active
demographic module.
We therefore isolate the task-common survey capability in a shared \tasklora{}, allowing each demographic module to focus on specific patterns.

Specifically, we sample survey items from the training split, remove respondent demographics, and replace each response with a valid option identifier sampled uniformly from the candidate set. 
The identifiers are balanced across the constructed dataset to avoid encoding demographic preference signals and systematic option-position bias. 
In this way, the \tasklora{} primarily learns the common survey
response protocol and valid output format.

Let $\mathcal{D}_{\mathrm{task}} = \left\{ \left(q_i,a_i^{\mathrm{rand}}\right) \right\}_{i=1}^{N_{\mathrm{task}}}$ denote the resulting training set, where $q_i$ contains a survey question and its candidate answers, and $a_i^{\mathrm{rand}}$ is the sampled option identifier. We optimize the \tasklora{} parameters $\phi_{\mathrm{task}}$ using \begin{equation} \mathcal{L}_{\mathrm{task}} = -\frac{1}{N_{\mathrm{task}}} \sum_{i=1}^{N_{\mathrm{task}}} \log p_{\theta_0,\phi_{\mathrm{task}}} \left( a_i^{\mathrm{rand}}\mid q_i \right), \label{eq:task_loss} \end{equation} where $\theta_0$ denotes the frozen backbone parameters. The loss is computed only on the target option identifier. After Stage~1, the module is frozen and included in every composed model during training and inference.

\subsection{Stage 2: Demographic LoRA Training}
\label{sec:stage2}
The demographic modules may interact unpredictably when composed for a complete profile. We therefore optimize the modules activated by each profile jointly, allowing them to learn compatible attribute-specific response shifts.
Let
$\mathcal{S}_{\mathrm{train}}=\{(q_i,g_i,a_i)\}_{i=1}^{N}$,
where $q_i$ contains a survey question and candidate options, $g_i$ is
the respondent profile, and $a_i\in\mathcal{A}_{q_i}$ is the answer. The backbone and the \tasklora{} learned in Stage~1 remain frozen.
For each record, only the \demolora{} modules selected by $g_i$ are
activated.

Using the composition defined in
\cref{eq:demographic_composition}
and \cref{eq:effective_weight}, the forward propagation of an adapted layer
$\ell$ is
\begin{equation}
\mathbf{u}_{\ell,i}
=
\widetilde{W}_{\ell}(g_i)\mathbf{h}_{\ell,i}.
\label{eq:stage2_forward}
\end{equation}
Consequently, gradients from record $i$ update only the active demographic
modules and their dimension-level weights.

\paragraph{Dimension-level weights.}
The additive approximation may not hold equally across demographic dimensions, and different LoRA modules may also produce updates with different numerical scales. Uniform addition may therefore over- or under-emphasize particular dimensions. 
We assign one learnable scalar $w_d$ to each dimension, initialize it to one, and apply it to all values within that dimension. These weights adjust the relative scales of different demographic updates without introducing profile-specific parameters.

\paragraph{Task-Demographic subspace separation.}
Stage~2 combines profile-level response supervision with soft orthogonality regularization to prevent redundant encoding of shared task information and encourage demographic modules to learn complementary patterns.
Specifically, for a training batch $b$, let $\mathcal{I}_b$ denote its sample indices
and let $g_b$ denote the demographic profile shared by its samples.
Let $\Theta(g_b)$ denote the composed model parameters induced by
  $\widetilde{W}_{\ell}(g_b)$. The response loss is
  \begin{equation}
  \mathcal{L}_{\mathrm{resp}}(g_b)
  =
  -\frac{1}{|\mathcal{I}_b|}
  \sum_{i\in\mathcal{I}_b}
  \log p_{\Theta(g_b)}(a_i\mid q_i).
  \label{eq:response_loss}
  \end{equation}

Although the \tasklora{} remains frozen during Stage~2, a demographic
adapter may still learn input-space directions that overlap with those already represented by the task adapter. We therefore introduce a soft orthogonality regularizer to discourage such overlap between LoRA matrices.
For the demographic adapters active under profile $g_b$, we define
  \begin{equation}
  \mathcal{L}_{\mathrm{orth}}(g_b)
  =
  \frac{1}{|\mathcal{L}_A|\,|g_b|}
  \sum_{\ell\in\mathcal{L}_A}
  \sum_{(d,v)\in g_b}
  \frac{
  \left\|A_{\ell,t}A_{\ell,d,v}^{\top}\right\|_F^2
  }{r^2}.
  \label{eq:soft_orthogonality}
  \end{equation}
  Here, $\mathcal{L}_A$ denotes the set of LoRA-adapted layers.
  Since $A_{\ell,t}$ is frozen during demographic-module training,
  gradients from this regularizer flow only through the demographic
  down-projection matrices $A_{\ell,d,v}$.

  The complete Stage~2 objective for batch $b$ is
  \begin{equation}
  \mathcal{L}_{\mathrm{demo}}(g_b)
  =
  \mathcal{L}_{\mathrm{resp}}(g_b)
  +
  \lambda\mathcal{L}_{\mathrm{orth}}(g_b).
  \label{eq:stage2_objective}
  \end{equation}
  We optimize the demographic LoRA parameters and dimension-level weights
  using \cref{eq:stage2_objective}, while keeping the backbone and
  \tasklora{} fixed. All modules activated by the same profile are
  optimized jointly from the same response loss, allowing each module to
  account for its co-occurring demographic attributes. Algorithm~\ref{alg:parade_training} summarizes the complete two-stage training pipeline of \method{}.

\begin{algorithm}[t]
\caption{Two-stage training pipeline of \method{}.}
\label{alg:parade_training}
\small
\begin{algorithmic}[1]
\Require Frozen backbone parameters $\theta_0$; training set
$\mathcal{S}_{\mathrm{train}}$; demographic value sets
$\{\mathcal{V}_d\}_{d\in\mathcal{D}}$; regularization coefficient $\lambda$
\Ensure Trained Task LoRA $\phi_{\mathrm{task}}$, Demographic LoRAs
$\{\phi_{d,v}\}$, and dimension-level weights $\{w_d\}$

\Statex \textbf{Stage 1: Task LoRA training}
\State Construct $\mathcal{D}_{\mathrm{task}}$ by removing demographics and
assigning balanced random valid option identifiers
\State Initialize Task LoRA $\phi_{\mathrm{task}}$
\While{Stage~1 has not converged}
    \State Sample a mini-batch
    $\mathcal{B}_{t}\subset\mathcal{D}_{\mathrm{task}}$
    \State Update $\phi_{\mathrm{task}}$ using
    \cref{eq:task_loss}
\EndWhile

\Statex \textbf{Stage 2: Demographic LoRA training}
\State Initialize $\phi_{d,v}$ for all
$d\in\mathcal{D}$ and $v\in\mathcal{V}_d$
\State Initialize $w_d\gets 1$ for all $d\in\mathcal{D}$
\While{Stage~2 has not converged}
    \State Sample a profile-homogeneous mini-batch $b$ with profile $g_b$
    \State Activate
    $\Phi(g_b)=\{\phi_{d,v}\mid(d,v)\in g_b\}$
    \State Compose $\widetilde{W}_{\ell}(g_b)$ using
    \cref{eq:demographic_composition,eq:effective_weight}
    \State Compute $\mathcal{L}_{\mathrm{demo}}(g_b)$ using
    \cref{eq:stage2_objective}
    \State Update $\Phi(g_b)$ and weights
    $\{w_d\mid(d,v)\in g_b\}$
\EndWhile

\State \Return $\phi_{\mathrm{task}}$, $\{\phi_{d,v}\}$, and $\{w_d\}$
\end{algorithmic}
\end{algorithm}

\subsection{Inference and Compositional Generalization}
\label{sec:combine}

Given a question $q$ and target profile $g$, we activate the \tasklora{} and the \demolora{} modules corresponding to the values in
$g$. The composed weights are constructed using
\cref{eq:demographic_composition}
and \cref{eq:effective_weight}. All model parameters and dimension-level
weights are fixed during inference, and the input contains only the survey
question and its candidate answers.

During inference, the composed model is instructed to output the selected option identifier first, which is then extracted using a deterministic parser. Each sampled respondent contributes one prediction per question, and these predictions are aggregated within the evaluation population and normalized to estimate the response distribution.

Because composition operates on individual attribute values rather than
complete profiles, \method{} can instantiate an unseen demographic
combination whenever its constituent values have been observed during
training. Let $\mathcal{V}_d$ denote the value set of demographic dimension
$d$ in one LoRA.
Covering all possible complete profiles with profile-specific adaptation
requires $\prod_{d\in\mathcal{D}}|\mathcal{V}_d|$ independently trained
LoRAs, yielding
$\mathcal{O}\!\left(\prod_d|\mathcal{V}_d|\right)$ parameters and optimization processes. In contrast, \method{}
learns one shared Task LoRA and
$\sum_{d\in\mathcal{D}}|\mathcal{V}_d|$ component-level Demographic LoRAs,
requiring only
$\mathcal{O}\!\left((1+\sum_d|\mathcal{V}_d|)\right)$
parameters under joint training. Thus, \method{} reduces the training and
storage overhead from multiplicative scaling to additive scaling, while supporting unseen combinations without
additional training.

\begin{table*}[t]
\centering
{%
\small
\renewcommand{\arraystretch}{1.12}
\setlength{\tabcolsep}{0.5mm}
\begin{tabular}{@{}llcccccccccc@{}}
\toprule
\textbf{Model}
& \textbf{Method}
& \multicolumn{5}{c}{\textbf{WVS}}
& \multicolumn{5}{c}{\textbf{SocioBench}} \\
\cmidrule(lr){3-7}
\cmidrule(lr){8-12}
&
& \textbf{Acc. $\uparrow$}
& \textbf{MAE $\downarrow$}
& \textbf{EMD $\downarrow$}
& \textbf{JS $\downarrow$}
& \textbf{KL $\downarrow$}
& \textbf{Acc. $\uparrow$}
& \textbf{MAE $\downarrow$}
& \textbf{EMD $\downarrow$}
& \textbf{JS $\downarrow$}
& \textbf{KL $\downarrow$} \\
\midrule

\multirow{7}{*}{\shortstack{Llama\\-3.1-8B-\\Instruct}}
& Vanilla
& 37.9\std{1.0}
& 0.330\std{0.006}
& 0.330\std{0.006}
& 0.458\std{0.008}
& 14.794\std{0.241}
& 32.0\std{0.3}
& 0.296\std{0.002}
& 0.290\std{0.002}
& 0.492\std{0.003}
& 16.141\std{0.083} \\

& Persona
& 37.4\std{0.9}
& 0.317\std{0.006}
& 0.266\std{0.006}
& 0.361\std{0.009}
& 9.621\std{0.615}
& 34.6\std{0.2}
& 0.265\std{0.001}
& \underline{0.216}\std{0.001}
& 0.319\std{0.004}
& 8.779\std{0.186} \\

& SimVBG
& 41.9\std{0.9}
& 0.331\std{0.008}
& 0.270\std{0.009}
& 0.281\std{0.012}
& 7.253\std{0.406}
& 31.1\std{0.4}
& 0.315\std{0.002}
& 0.239\std{0.001}
& \underline{0.286}\std{0.004}
& \textbf{5.511}\std{0.234} \\

& LoRA
& 48.1\std{0.9}
& \underline{0.298}\std{0.004}
& 0.288\std{0.004}
& 0.334\std{0.008}
& 11.062\std{0.299}
& 37.3\std{0.2}
& 0.269\std{0.001}
& 0.262\std{0.001}
& 0.432\std{0.002}
& 14.694\std{0.070} \\

& MixLoRA
& \underline{48.3}\std{0.7}
& 0.299\std{0.004}
& 0.284\std{0.003}
& 0.315\std{0.005}
& 10.465\std{0.178}
& \textbf{38.2}\std{0.2}
& \textbf{0.262}\std{0.001}
& 0.251\std{0.001}
& 0.408\std{0.002}
& 13.898\std{0.059} \\

& CuMA
& 46.6\std{0.5}
& 0.311\std{0.004}
& \underline{0.248}\std{0.005}
& \underline{0.247}\std{0.005}
& \underline{7.106}\std{0.276}
& \underline{38.1}\std{0.2}
& 0.265\std{0.001}
& 0.219\std{0.001}
& 0.320\std{0.002}
& 9.163\std{0.089} \\

\rowcolor{paradeyellow}
\cellcolor{white}
& \textbf{\method{}}
& \textbf{48.4}\std{0.9}
& \textbf{0.284}\std{0.004}
& \textbf{0.172}\std{0.004}
& \textbf{0.181}\std{0.006}
& \textbf{4.943}\std{0.237}
& 38.0\std{0.3}
& \underline{0.265}\std{0.001}
& \textbf{0.189}\std{0.001}
& \textbf{0.244}\std{0.002}
& \underline{6.145}\std{0.076} \\

\midrule

\multirow{7}{*}{\shortstack{Qwen\\3.5-9B}}
& Vanilla
& 40.2\std{0.4}
& 0.344\std{0.003}
& 0.344\std{0.003}
& 0.447\std{0.006}
& 14.050\std{0.081}
& 32.2\std{0.3}
& 0.301\std{0.002}
& 0.294\std{0.002}
& 0.491\std{0.004}
& 16.053\std{0.084} \\

& Persona
& 40.3\std{0.5}
& 0.323\std{0.005}
& 0.272\std{0.006}
& 0.336\std{0.008}
& 8.191\std{0.252}
& 33.8\std{0.2}
& 0.272\std{0.001}
& 0.224\std{0.001}
& 0.357\std{0.003}
& 10.119\std{0.184} \\

& SimVBG
& 43.4\std{1.2}
& 0.304\std{0.007}
& \underline{0.220}\std{0.007}
& \underline{0.224}\std{0.008}
& \underline{5.764}\std{0.313}
& 35.5\std{0.2}
& 0.265\std{0.001}
& \underline{0.183}\std{0.001}
& \underline{0.233}\std{0.002}
& \underline{4.595}\std{0.150} \\

& LoRA
& 46.8\std{1.1}
& 0.320\std{0.005}
& 0.318\std{0.005}
& 0.365\std{0.011}
& 12.283\std{0.274}
& \underline{37.6}\std{0.2}
& 0.268\std{0.001}
& 0.254\std{0.001}
& 0.425\std{0.002}
& 14.450\std{0.051} \\

& MixLoRA
& 47.7\std{1.0}
& 0.310\std{0.006}
& 0.304\std{0.005}
& 0.347\std{0.010}
& 11.622\std{0.261}
& 36.9\std{0.3}
& 0.277\std{0.002}
& 0.259\std{0.002}
& 0.419\std{0.003}
& 14.277\std{0.076} \\

& CuMA
& \underline{47.9}\std{0.9}
& \underline{0.299}\std{0.005}
& 0.234\std{0.005}
& 0.249\std{0.011}
& 6.723\std{0.409}
& 37.6\std{0.3}
& \underline{0.263}\std{0.001}
& 0.221\std{0.002}
& 0.316\std{0.002}
& 8.866\std{0.166} \\

\rowcolor{paradeyellow}
\cellcolor{white}
& \textbf{\method{}}
& \textbf{48.9}\std{0.8}
& \textbf{0.289}\std{0.003}
& \textbf{0.176}\std{0.004}
& \textbf{0.195}\std{0.003}
& \textbf{5.188}\std{0.280}
& \textbf{38.8}\std{0.3}
& \textbf{0.254}\std{0.001}
& \textbf{0.167}\std{0.001}
& \textbf{0.196}\std{0.003}
& \textbf{4.385}\std{0.102} \\

\midrule

\multirow{7}{*}{\shortstack{Ministral\\-3-8B-\\Instruct}}
& Vanilla
& 37.0\std{0.3}
& 0.332\std{0.004}
& 0.332\std{0.004}
& 0.479\std{0.002}
& 15.063\std{0.069}
& 32.2\std{0.3}
& 0.283\std{0.001}
& 0.269\std{0.002}
& 0.484\std{0.003}
& 15.987\std{0.080} \\

& Persona
& 38.9\std{0.7}
& 0.302\std{0.006}
& 0.233\std{0.006}
& 0.296\std{0.005}
& 7.417\std{0.236}
& 33.7\std{0.2}
& 0.264\std{0.001}
& 0.207\std{0.002}
& 0.339\std{0.004}
& 9.603\std{0.100} \\

& SimVBG
& 37.8\std{1.0}
& 0.323\std{0.006}
& 0.224\std{0.005}
& 0.234\std{0.006}
& \underline{4.959}\std{0.399}
& 34.2\std{0.1}
& 0.278\std{0.002}
& \textbf{0.163}\std{0.001}
& \textbf{0.173}\std{0.002}
& \textbf{2.888}\std{0.199} \\

& LoRA
& 47.9\std{0.7}
& 0.308\std{0.003}
& 0.306\std{0.003}
& 0.353\std{0.006}
& 11.971\std{0.181}
& 37.6\std{0.3}
& 0.271\std{0.001}
& 0.269\std{0.001}
& 0.431\std{0.003}
& 14.630\std{0.073} \\

& MixLoRA
& \underline{48.6}\std{1.0}
& 0.303\std{0.004}
& 0.281\std{0.006}
& 0.326\std{0.009}
& 10.972\std{0.243}
& 36.8\std{0.2}
& 0.278\std{0.001}
& 0.259\std{0.001}
& 0.421\std{0.001}
& 14.245\std{0.044} \\

& CuMA
& \textbf{49.4}\std{0.9}
& \underline{0.297}\std{0.004}
& \underline{0.208}\std{0.002}
& \underline{0.204}\std{0.003}
& 5.685\std{0.212}
& \textbf{38.7}\std{0.1}
& \textbf{0.256}\std{0.001}
& 0.206\std{0.001}
& 0.286\std{0.002}
& 7.864\std{0.091} \\

\rowcolor{paradeyellow}
\cellcolor{white}
& \textbf{\method{}}
& 47.7\std{0.9}
& \textbf{0.293}\std{0.007}
& \textbf{0.131}\std{0.007}
& \textbf{0.115}\std{0.006}
& \textbf{2.339}\std{0.116}
& \underline{37.9}\std{0.4}
& \underline{0.261}\std{0.002}
& \underline{0.172}\std{0.002}
& \underline{0.200}\std{0.004}
& \underline{4.016}\std{0.136} \\

\bottomrule
\end{tabular}
}%
\caption{Main results on WVS and SocioBench. Mean and sample standard deviation are reported across five test-user resampling seeds. Accuracy is reported in percent. Bold and underlined values denote the best and second-best mean results, respectively.}
\label{tab:main_results}
\end{table*}

\begin{table*}[t]
\centering
{%
\small
\renewcommand{\arraystretch}{1.12}
\setlength{\tabcolsep}{0.5mm}
\begin{tabular}{@{}llcccccccccc@{}}
\toprule
\textbf{Model}
& \textbf{Method}
& \multicolumn{5}{c}{\textbf{WVS}}
& \multicolumn{5}{c}{\textbf{SocioBench}} \\
\cmidrule(lr){3-7}
\cmidrule(lr){8-12}
&
& \textbf{Acc. $\uparrow$}
& \textbf{MAE $\downarrow$}
& \textbf{EMD $\downarrow$}
& \textbf{JS $\downarrow$}
& \textbf{KL $\downarrow$}
& \textbf{Acc. $\uparrow$}
& \textbf{MAE $\downarrow$}
& \textbf{EMD $\downarrow$}
& \textbf{JS $\downarrow$}
& \textbf{KL $\downarrow$} \\
\midrule

\multirow{7}{*}{\shortstack{Llama\\-3.1-8B-\\Instruct}}
& Vanilla
& 37.8\std{0.2}
& 0.328\std{0.002}
& 0.327\std{0.002}
& 0.459\std{0.002}
& 14.781\std{0.035}
& 32.0\std{0.9}
& 0.295\std{0.009}
& 0.288\std{0.010}
& 0.513\std{0.013}
& 16.377\std{0.231} \\

& Persona
& 37.5\std{0.4}
& 0.315\std{0.006}
& 0.265\std{0.005}
& 0.362\std{0.006}
& 10.117\std{0.359}
& 34.8\std{1.0}
& 0.266\std{0.008}
& \underline{0.225}\std{0.008}
& 0.368\std{0.012}
& 11.020\std{0.236} \\

& SimVBG
& 42.4\std{0.5}
& 0.328\std{0.005}
& 0.269\std{0.006}
& 0.279\std{0.005}
& 7.250\std{0.181}
& 30.8\std{1.2}
& 0.317\std{0.009}
& 0.258\std{0.011}
& 0.374\std{0.020}
& \underline{10.288}\std{0.511} \\

& LoRA
& 47.2\std{0.8}
& 0.309\std{0.005}
& 0.303\std{0.005}
& 0.357\std{0.007}
& 12.131\std{0.261}
& 36.9\std{0.8}
& 0.270\std{0.007}
& 0.264\std{0.007}
& 0.454\std{0.006}
& 15.017\std{0.218} \\

& MixLoRA
& \underline{47.5}\std{0.5}
& \underline{0.304}\std{0.005}
& 0.281\std{0.004}
& 0.313\std{0.004}
& 10.349\std{0.152}
& 37.4\std{1.5}
& 0.266\std{0.007}
& 0.255\std{0.007}
& 0.436\std{0.010}
& 14.455\std{0.280} \\

& CuMA
& 46.3\std{0.7}
& 0.312\std{0.007}
& \underline{0.251}\std{0.006}
& \underline{0.248}\std{0.004}
& \underline{7.158}\std{0.148}
& \underline{38.0}\std{0.5}
& \underline{0.265}\std{0.005}
& 0.228\std{0.008}
& \underline{0.368}\std{0.012}
& 11.674\std{0.268} \\

\rowcolor{paradeyellow}
\cellcolor{white}
& \textbf{\method{}}
& \textbf{47.6}\std{0.8}
& \textbf{0.288}\std{0.006}
& \textbf{0.169}\std{0.003}
& \textbf{0.182}\std{0.004}
& \textbf{4.996}\std{0.222}
& \textbf{38.3}\std{1.2}
& \textbf{0.264}\std{0.008}
& \textbf{0.200}\std{0.007}
& \textbf{0.301}\std{0.014}
& \textbf{9.364}\std{0.615} \\

\midrule

\multirow{7}{*}{\shortstack{Qwen\\3.5-9B}}
& Vanilla
& 39.8\std{0.6}
& 0.344\std{0.004}
& 0.344\std{0.005}
& 0.449\std{0.007}
& 14.151\std{0.076}
& 31.6\std{1.6}
& 0.302\std{0.006}
& 0.296\std{0.007}
& 0.518\std{0.018}
& 16.508\std{0.441} \\

& Persona
& 39.6\std{0.5}
& 0.326\std{0.007}
& 0.274\std{0.007}
& 0.340\std{0.006}
& 8.262\std{0.189}
& 33.1\std{1.3}
& 0.274\std{0.003}
& 0.234\std{0.002}
& 0.415\std{0.014}
& 12.659\std{0.353} \\

& SimVBG
& 43.2\std{0.7}
& 0.305\std{0.007}
& \underline{0.217}\std{0.005}
& \underline{0.219}\std{0.008}
& \underline{5.684}\std{0.264}
& 36.3\std{1.8}
& \underline{0.262}\std{0.010}
& \underline{0.202}\std{0.006}
& \underline{0.310}\std{0.008}
& \underline{8.881}\std{0.235} \\

& LoRA
& 46.2\std{0.4}
& 0.324\std{0.004}
& 0.322\std{0.004}
& 0.370\std{0.004}
& 12.415\std{0.137}
& \underline{37.8}\std{0.4}
& 0.265\std{0.004}
& 0.252\std{0.004}
& 0.443\std{0.002}
& 14.714\std{0.072} \\

& MixLoRA
& 46.9\std{0.6}
& 0.316\std{0.005}
& 0.304\std{0.006}
& 0.350\std{0.004}
& 11.754\std{0.141}
& 36.7\std{0.7}
& 0.277\std{0.007}
& 0.259\std{0.008}
& 0.440\std{0.007}
& 14.506\std{0.178} \\

& CuMA
& \underline{47.7}\std{0.5}
& \underline{0.300}\std{0.006}
& 0.233\std{0.007}
& 0.247\std{0.005}
& 6.695\std{0.234}
& 37.7\std{0.7}
& 0.264\std{0.006}
& 0.228\std{0.007}
& 0.362\std{0.008}
& 11.308\std{0.396} \\

\rowcolor{paradeyellow}
\cellcolor{white}
& \textbf{\method{}}
& \textbf{48.3}\std{0.7}
& \textbf{0.291}\std{0.008}
& \textbf{0.172}\std{0.009}
& \textbf{0.199}\std{0.006}
& \textbf{5.376}\std{0.085}
& \textbf{39.3}\std{1.2}
& \textbf{0.250}\std{0.006}
& \textbf{0.183}\std{0.007}
& \textbf{0.267}\std{0.008}
& \textbf{7.879}\std{0.389} \\

\midrule

\multirow{7}{*}{\shortstack{Ministral\\-3-8B-\\Instruct}}
& Vanilla
& 36.5\std{0.3}
& 0.332\std{0.003}
& 0.331\std{0.003}
& 0.483\std{0.004}
& 15.143\std{0.052}
& 31.9\std{0.9}
& 0.284\std{0.006}
& 0.270\std{0.007}
& 0.508\std{0.007}
& 16.314\std{0.237} \\

& Persona
& 39.1\std{0.5}
& 0.300\std{0.006}
& 0.235\std{0.003}
& 0.296\std{0.006}
& 7.402\std{0.149}
& 33.0\std{1.2}
& \underline{0.267}\std{0.003}
& 0.217\std{0.007}
& 0.386\std{0.008}
& 11.803\std{0.270} \\

& SimVBG
& 38.2\std{0.4}
& 0.320\std{0.007}
& 0.224\std{0.005}
& 0.227\std{0.006}
& \underline{4.630}\std{0.244}
& 33.8\std{1.8}
& 0.278\std{0.014}
& \textbf{0.191}\std{0.002}
& \textbf{0.272}\std{0.008}
& \textbf{7.224}\std{0.360} \\

& LoRA
& 47.0\std{0.4}
& 0.314\std{0.004}
& 0.313\std{0.004}
& 0.366\std{0.004}
& 12.351\std{0.149}
& 36.9\std{1.7}
& 0.273\std{0.009}
& 0.270\std{0.009}
& 0.454\std{0.014}
& 15.078\std{0.425} \\

& MixLoRA
& \underline{48.2}\std{0.2}
& 0.305\std{0.004}
& 0.284\std{0.003}
& 0.331\std{0.002}
& 11.121\std{0.088}
& 35.9\std{1.4}
& 0.282\std{0.008}
& 0.265\std{0.007}
& 0.449\std{0.010}
& 14.757\std{0.306} \\

& CuMA
& \textbf{49.2}\std{0.4}
& \underline{0.299}\std{0.005}
& \underline{0.213}\std{0.005}
& \underline{0.208}\std{0.004}
& 5.840\std{0.214}
& \textbf{38.3}\std{0.9}
& \textbf{0.256}\std{0.009}
& 0.216\std{0.007}
& 0.340\std{0.006}
& 10.514\std{0.309} \\

\rowcolor{paradeyellow}
\cellcolor{white}
& \textbf{\method{}}
& 47.5\std{0.6}
& \textbf{0.297}\std{0.006}
& \textbf{0.131}\std{0.007}
& \textbf{0.119}\std{0.002}
& \textbf{2.448}\std{0.152}
& \underline{37.1}\std{1.4}
& 0.268\std{0.011}
& \underline{0.202}\std{0.010}
& \underline{0.298}\std{0.017}
& \underline{8.724}\std{0.608} \\

\bottomrule
\end{tabular}
}%
\caption{Results on unseen combinations. Mean and sample standard deviation are reported across five test-user resampling seeds. Accuracy is reported in percent. Bold and underlined values denote the best and second-best mean results, respectively.}
\label{tab:unseen_combinations}
\end{table*}

\section{Experimental Setup}

\subsection{Datasets and Metrics}

We conduct experiments on two survey datasets: \textbf{World Values Survey (WVS) Wave~7} \citep{haerpfer2022world} and \textbf{SocioBench} \citep{wang2025sociobench}. Both datasets cover diverse survey topics and provide demographic annotations of respondents. We consider eight demographic dimensions---gender, age group, country, religion, education, marital status, employment status, and urban/rural residence---which capture demographic backgrounds from complementary perspectives. These attributes are selected based on their mutual information with responses and coverage across datasets. 

We evaluate individual responses using \textbf{Accuracy} and normalized \textbf{Mean Absolute Error (MAE)}, where MAE is divided by $K_q-1$ for a question with $K_q$ options. At the group level, we compare empirical and simulated response distributions using \textbf{Earth Mover's Distance (EMD)}, \textbf{Jensen-Shannon (JS) divergence}, and \textbf{Kullback-Leibler (KL) divergence}. EMD accounts for the ordinal distances between options, while JS and KL measure distributional discrepancy. 
Due to space constraints, dataset splits, demographic details, and definitions of metrics are provided in Appendix~B.

\subsection{Baseline Methods}

We compare \method{} with six baselines. \textbf{Vanilla Baseline} uses the frozen backbone and receives only the survey questions. \textbf{Persona Prompting} conditions the model on demographic attributes provided in the system prompt \citep{lutz2025prompt}. \textbf{SimVBG} converts demographic information into a backstory and uses a cognitive--affective--behavioral process to simulate values \citep{du2025simvbg}. \textbf{LoRA} trains a shared low-rank adaptation on all records \citep{hu2022lora}. \textbf{MixLoRA} learns multiple LoRA experts and routes each input among them without demographic conditioning \citep{li2024mixlora}. \textbf{CuMA} performs demographic-aware routing over specialized LoRA experts to model value variation across population groups \citep{sun2026cuma}. All methods use the same backbones, data partitions, answer format, and evaluation protocol. Detailed prompts and implementation settings are provided in Appendix~B.




\subsection{Implementation Details}

We conduct experiments with three backbones: Llama-3.1-8B-Instruct, Qwen3.5-9B, and Ministral-3-8B-Instruct. LoRA modules are applied to the query, key, and value projections in self-attention, as well as the gate, up, and down projections in the MLP. 
All tunable hyperparameters are selected on the validation set.
The complete hyperparameter settings are provided in Appendix~B.

\section{Experiments and Results}

We structure our experiments around four research questions: whether \method{} outperforms baseline methods on value simulation (\textbf{RQ1}); whether recombining demographic modules enables generalization to unseen profiles (\textbf{RQ2}); how each component contributes to performance (\textbf{RQ3}); and whether \method{} preserves population-level heterogeneity rather than concentrating predictions on a few options (\textbf{RQ4}).

To improve evaluation reliability, we conduct five evaluation runs. In each run, we resample the test respondents using a different random seed. For trainable methods, we use the same trained checkpoint across all five runs. The tables report the mean and standard deviation across the five runs, while the results of each run are provided in Appendix~C.

\subsection{Main Results}

To answer \textbf{RQ1}, we compare \method{} with baselines in terms of both individual-level prediction quality and group-level distributional fidelity. Table~\ref{tab:main_results} summarizes the results. Overall, \method{} achieves the best or second-best result in 28 of the 30 backbone--dataset--metric combinations, showing strong performance across settings.

On WVS, it reduces EMD over the
strongest baseline by 30.6\%, 20.0\%, and 37.0\% with
Llama-3.1-8B-Instruct, Qwen3.5-9B, and Ministral-3-8B-Instruct,
respectively. With Ministral-3-8B-Instruct, JS and KL are reduced by
43.6\% and 52.8\%. On SocioBench with Qwen3.5-9B, \method{} is best on all metrics, improving Accuracy by 1.2 percentage points while reducing
EMD, JS, and KL by 8.7\%, 15.9\%, and 4.6\%. These results show that
\method{} substantially improves population-level distribution alignment
while preserving competitive individual-level prediction quality. 

\subsection{Generalization to Unseen Combinations}

To answer \textbf{RQ2}, we evaluate complete demographic
combinations excluded from training. Table~\ref{tab:unseen_combinations} shows that \method{} ranks first in 24 settings and second in four others.

On WVS, it reduces EMD, JS, and KL relative to the strongest baselines by 32.7\%,
26.6\%, and 30.2\% with Llama-3.1-8B-Instruct, and by 38.5\%, 42.8\%, and
47.1\% with Ministral-3-8B-Instruct. On SocioBench with Qwen3.5-9B,
\method{} improves Accuracy by 1.5 percentage points and reduces EMD, JS,
and KL by 9.4\%, 13.9\%, and 11.3\%, respectively. Since no adapter is trained for these complete profiles, the results demonstrate that the learned modules capture
reusable attribute-level effects that can be recombined to construct
previously unseen populations.

\subsection{Ablation Studies}

To answer \textbf{RQ3}, we retrain four variants that remove the Task LoRA, dimension-level weights,
orthogonality regularization, or Demographic LoRAs under identical settings. 

Table~\ref{tab:ablation_wvs_ministral} reports ablations with Ministral-3-8B-Instruct. The full model achieves the best MAE, EMD, JS, and KL. Removing the Demographic LoRAs causes the largest degradation: Accuracy drops by 20.72 percentage points, while EMD, JS, and KL become 4.1, 5.0, and 7.1 times those of the
full model. Removing orthogonality regularization increases these distributional errors by 55.4\%, 63.7\%, and 79.0\%, while fixing weights increases them by 24.8\%, 37.9\%, and 51.1\%.
Removing the Task LoRA increases EMD, JS, and KL by 32.1\%,
48.4\%, and 73.7\%. These results support the complementary roles of attribute-level demographic adaptation, task--demographic separation, and composition calibration in recovering population-level distributions. Sensitivity analysis of $\lambda$ in Eq.~\eqref{eq:stage2_objective} is provided in Appendix~C.

\begin{table}[t]
\centering
{%
\small
\renewcommand{\arraystretch}{1.12}
\setlength{\tabcolsep}{0.5mm}
\begin{tabular*}{\columnwidth}{@{}l@{\extracolsep{\fill}}ccccc@{}}
\toprule
\textbf{Setting}
& \textbf{Acc. $\uparrow$}
& \textbf{MAE $\downarrow$}
& \textbf{EMD $\downarrow$}
& \textbf{JS $\downarrow$}
& \textbf{KL $\downarrow$} \\
\midrule

\textbf{Full}
& 47.71
& \textbf{0.2934}
& \textbf{0.1306}
& \textbf{0.1150}
& \textbf{2.3390} \\

\midrule

w/o Task
& \textbf{48.31}
& 0.2940
& 0.1725
& 0.1707
& 4.0631 \\

w/o Weight
& 47.28
& 0.2997
& 0.1630
& 0.1586
& 3.5333 \\

w/o Orth.
& 46.20
& 0.3119
& 0.2030
& 0.1883
& 4.1870 \\

w/o Demo.
& 26.99
& 0.5510
& 0.5404
& 0.5792
& 16.5265 \\

\bottomrule
\end{tabular*}
}%
\caption{Ablation results on WVS using Ministral-3-8B-Instruct.
Orth. denotes orthogonality regularization, while Demo. denotes Demographic
LoRAs. Accuracy is reported in percent. Bold values denote the best results.}
\label{tab:ablation_wvs_ministral}
\end{table}

\subsection{Case Study}

\begin{figure}[t]
    \centering
    \includegraphics[width=\columnwidth]{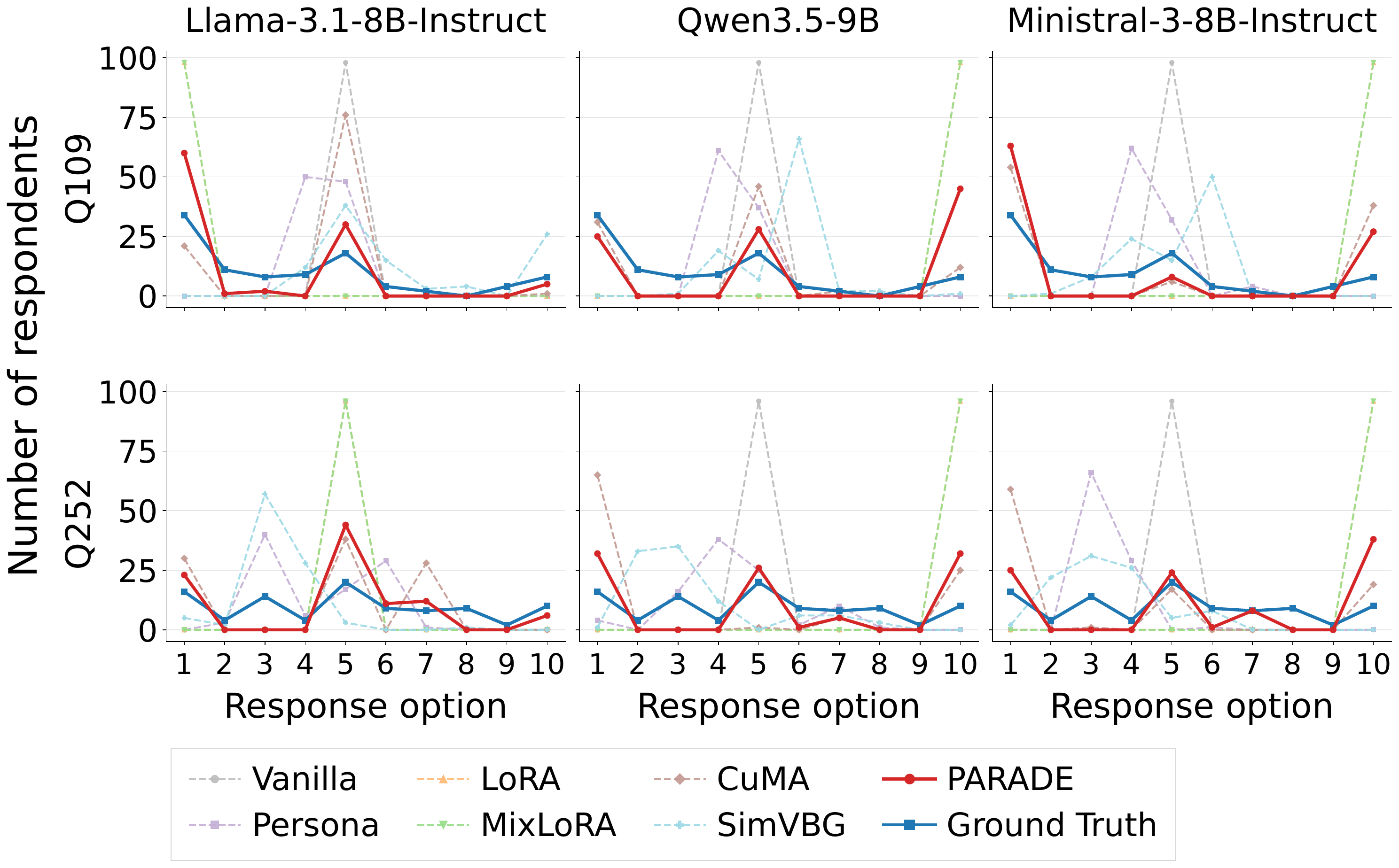}
    \caption{
        Response distributions on two WVS questions. Rows correspond to Q109 (Views on competition; 1 = Competition is good; 10 = Competition is harmful) and Q252 (Satisfaction with the political system; 1 = Not satisfied at all; 10 = Completely satisfied), respectively. Each curve shows the number of respondents assigned to each response option by a prediction method or observed in the ground-truth data.
    }
    \label{fig:case_study}
\end{figure}

To answer \textbf{RQ4}, we examine whether each method recovers both dominant preferences and variation across response options. ~\cref{fig:case_study} compares the predicted and empirical response distributions for WVS questions Q109 (Views on competition) and Q252 (Satisfaction with the political system).

Several baselines concentrate their predictions on only a few choices, thereby underestimating the dispersion observed among human respondents. In contrast, \method{} better captures the major modes, option frequencies, and overall spread of the empirical distributions, providing qualitative evidence that it better reflects population-level response heterogeneity.

\section{Related Work}

\subsection{LLM Social Simulation and Prompting}

LLMs have been used as synthetic respondents for social experiments, opinion prediction, and interaction studies \citep{argyle2023out,aher2022using,kim2023ai,ashokkumar2026large}. They can capture aggregate and group-specific response patterns \citep{santurkar2023whose,durmus2023towards,cao2025specializing,meister2025benchmarking}, but default responses may misrepresent some groups and flatten variation across them \citep{dominguez2024questioning,ma2024potential,santurkar2023whose,wang2025large}. We therefore study group-level value simulation conditioned on multidimensional demographic profiles.

Most methods condition LLMs through personas, generated backstories, or few-shot examples \citep{sun2024random,du2025simvbg,du2026twinvoice,li2024steerability,cho2024llm}. Although training-free and compatible with black-box models, these methods are sensitive to persona wording, prone to identity drift and stereotypical responses, and incur additional inference cost \citep{lutz2025prompt,hu2024quantifying,sun2025sociodemographic,cheng2023marked}. These drawbacks motivate efficient and data-grounded control.

\subsection{Parameter Adaptation and Injection}

Parameter-based methods learn human behavior and values through supervised fine-tuning or preference optimization  \citep{kolluri2025finetuning,tan2024democratizing,zhao2023group,huang2026distribution}. Survey-specific training improves agreement with empirical distributions \citep{cao2025specializing,suh2025language}, while adapter routing and hidden-state interventions encode cultural or social identities \citep{sun2026cuma,wang2026parametric}. These studies establish the effectiveness of parameter-space control.

Low-rank adaptation (LoRA) enables lightweight parameter adaptation, while subsequent methods support routing, composition, and dynamic weighting of multiple adapters or LoRAs \citep{hu2022lora,li2024mixlora,pfeiffer2021adapterfusion,huang2023lorahub,wang2024loraflow,wang2023orthogonal}. Parametric Retrieval-Augmented Generation (PRAG) encodes external information into reusable parameters, while its decoupled variant separates shared task behavior from specific knowledge to facilitate composition \citep{su2025parametric,su2025dynamic,su2026retrievable,su2026decoupling,tang2025role}. Following this paradigm, \method{} separates survey-solving ability from demographic value patterns and composes demographic LoRAs for each profile.

\section{Conclusion}

We introduced \textbf{\method{}}, a composable parametric demographic
injection framework for population-specific value simulation. 
\method{} separates shared survey capabilities from demographic-conditioned value patterns, 
reduces task--demographic overlap, and calibrates module contributions during composition. 
This design constructs target population models without repeated demographic prompting or separate adapters for complete profiles.
Experiments on WVS and SocioBench across three backbones demonstrate
improved performance and strong generalization to unseen demographic
combinations. Future work could incorporate interaction-aware composition to capture non-additive demographic effects and question-conditioned calibration to adapt module contributions across survey questions.

\section*{Ethical Statement}

Synthetic respondents should complement rather than replace representative surveys, as demographic conditioning may reproduce stereotypes and is unsuitable for inferring individual beliefs or supporting high-stakes decisions. We use anonymized survey data in accordance with dataset licenses and make no causal claims about the learned parameters.

\bibliography{references}

\clearpage

\appendix

\section{Detailed Derivation of Additive Demographic Composition}
\label{app:theory}

Additive composition can be motivated under a conditional-independence
approximation. We derive the probability- and logit-space result step by
step, connect it to LoRA addition at an adapted linear layer, and clarify the
scope and limitations of this main-effect interpretation.

\subsection{Probability- and Logit-Space Derivation}

Let $q$ denote a survey question with candidate answers, $a$ an answer option,
and $\mathbf v=(v_1,\ldots,v_m)$ the demographic values of a respondent. We write $p_0(a\mid q)$ for the demographic-agnostic response distribution
of the pretrained backbone, $p_i(a\mid q,v_i)$ for the distribution
conditioned on value $v_i$, and $p(a\mid q,\mathbf v)$ for the
complete-profile distribution.

Assume that the demographic dimensions are conditionally independent given
the question and answer:
\begin{equation}
p(\mathbf v\mid a,q)
\approx \prod_{i=1}^{m}p(v_i\mid a,q).
\label{eq:cond_ind}
\end{equation}
This factorization is a working main-effect approximation used only to
motivate the composition rule. It need not hold exactly in survey data;
residual dependence among demographic dimensions corresponds to interaction
effects that are not explicitly represented by the derivation below.
Bayes' rule first gives
\begin{equation}
p(a\mid q,\mathbf v)
=\frac{p(\mathbf v\mid a,q)p_0(a\mid q)}{p(\mathbf v\mid q)}.
\label{eq:joint_bayes}
\end{equation}
For each demographic dimension, applying Bayes' rule separately gives
\begin{align}
p_i(a\mid q,v_i)
&=\frac{p(v_i\mid a,q)p_0(a\mid q)}{p(v_i\mid q)},\\
p(v_i\mid a,q)
&=\frac{p_i(a\mid q,v_i)p(v_i\mid q)}{p_0(a\mid q)}.
\label{eq:single_bayes}
\end{align}
Substituting Equation~\ref{eq:single_bayes} into the conditional-independence
assumption yields
\begin{align}
p(\mathbf v\mid a,q)
&\approx \prod_{i=1}^{m}
\frac{p_i(a\mid q,v_i)p(v_i\mid q)}{p_0(a\mid q)}\\
&=\frac{\prod_{i=1}^{m}p_i(a\mid q,v_i)
\prod_{i=1}^{m}p(v_i\mid q)}{p_0(a\mid q)^m}.
\label{eq:expanded_conditional}
\end{align}
Inserting this expression into Equation~\ref{eq:joint_bayes} gives
\begin{align}
p(a\mid q,\mathbf v)
&\approx
\frac{p_0(a\mid q)}{p(\mathbf v\mid q)}
\frac{\prod_{i=1}^{m}p_i(a\mid q,v_i)
\prod_{i=1}^{m}p(v_i\mid q)}{p_0(a\mid q)^m}\\
&=\frac{\prod_{i=1}^{m}p_i(a\mid q,v_i)}
{p_0(a\mid q)^{m-1}}
\frac{\prod_{i=1}^{m}p(v_i\mid q)}{p(\mathbf v\mid q)}\\
&=C(q,\mathbf v)
\frac{\prod_{i=1}^{m}p_i(a\mid q,v_i)}
{p_0(a\mid q)^{m-1}},
\label{eq:product_experts}
\end{align}
where
$C(q,\mathbf v)=\prod_i p(v_i\mid q)/p(\mathbf v\mid q)$ does not depend on
the answer option $a$. Thus, each single-attribute conditional contributes
one factor, while division by $p_0^{m-1}$ removes the repeatedly counted base
distribution.

We next show the corresponding logit relation without skipping the
normalization terms. Let $s_0(a,q)$, $s_i(a,q,v_i)$, and
$s_g(a,q,\mathbf v)$ denote logits for the three distributions, with
normalizers $Z_0$, $Z_i$, and $Z_g$. For example,
\begin{equation}
\log p_i(a\mid q,v_i)=s_i(a,q,v_i)-\log Z_i(q,v_i).
\label{eq:logit_probability}
\end{equation}
Taking logarithms of Equation~\ref{eq:product_experts} and substituting the
softmax identities gives
\begin{equation}
\begin{aligned}
s_g(a,q,\mathbf v)-\log Z_g
&\approx \sum_{i=1}^{m}
\left[s_i(a,q,v_i)-\log Z_i\right]\\
&\quad -(m-1)\left[s_0(a,q)-\log Z_0\right]\\
&\quad +\log C(q,\mathbf v)\\[2pt]
&=\sum_{i=1}^{m}s_i(a,q,v_i)-(m-1)s_0(a,q)\\
&\quad +K(q,\mathbf v).
\end{aligned}
\label{eq:expanded_logits}
\end{equation}
where
\begin{equation}
K(q,\mathbf v)=
-\sum_{i=1}^{m}\log Z_i+(m-1)\log Z_0+\log C(q,\mathbf v)
\end{equation}
is independent of $a$. Moving $\log Z_g$ into the same answer-independent
constant therefore gives
\begin{equation}
s_g(a,q,\mathbf v)
\doteq \sum_{i=1}^{m}s_i(a,q,v_i)-(m-1)s_0(a,q),
\label{eq:joint_logits}
\end{equation}
where $\doteq$ denotes equality up to an additive constant shared by all
answer options. Subtracting $s_0(a,q)$ from both sides and regrouping terms
gives the desired additive shift:
\begin{align}
s_g(a,q,\mathbf v)-s_0(a,q)
&\doteq \sum_{i=1}^{m}s_i(a,q,v_i)-m s_0(a,q)\\
&=\sum_{i=1}^{m}
\left[s_i(a,q,v_i)-s_0(a,q)\right].
\label{eq:additive_logit_shift}
\end{align}
Because softmax is invariant to a shared logit offset, the constant represented
by $\doteq$ does not change the response distribution. This result provides a
probabilistic motivation for additive main effects; it does not by itself
identify how separately learned parameter modules must decompose those effects.

\subsection{From LoRA Addition to Additive Shifts}

Consider one adapted linear layer with shared demographic-independent weight
$W_T=W+\Delta W_t$, where $W$ is the frozen backbone weight and
$\Delta W_t$ is the frozen Task LoRA update. For an input $h$, the output
without a Demographic LoRA is
\begin{equation}
o_T=W_T h.
\end{equation}
For demographic value $v_i$, let
$\Delta W_i(v_i)=(\alpha/r)B_i(v_i)A_i(v_i)$ be its LoRA update. Activating
this module gives
\begin{align}
o_i
&=\left(W_T+\Delta W_i(v_i)\right)h\\
&=W_T h+\Delta W_i(v_i)h\\
&=o_T+\Delta W_i(v_i)h.
\label{eq:single_layer_shift}
\end{align}
Activating all modules in the profile gives
\begin{align}
o_g
&=\left(W_T+\sum_{i=1}^{m}\Delta W_i(v_i)\right)h\\
&=W_T h+\sum_{i=1}^{m}\Delta W_i(v_i)h\\
&=o_T+\sum_{i=1}^{m}\left(o_i-o_T\right).
\label{eq:layer_addition}
\end{align}
Thus, LoRA updates and their layer-output shifts add exactly at an adapted
linear layer for a fixed input $h$. This identity does not imply exact
additivity for the complete network, because active modules can change hidden
states that subsequently pass through nonlinear layers.

Let $s_T$ denote the logits produced by the frozen backbone and Task LoRA,
and let $\Delta s_i$ denote the network-level logit perturbation associated
with activating value-specific module $i$ relative to this shared reference.
We use the local-additivity approximation
\begin{equation}
s_{\mathrm{comp}}(a,q,\mathbf v)-s_T(a,q)
\approx\sum_{i=1}^{m}\Delta s_i(a,q,v_i).
\label{eq:parameter_logit_link}
\end{equation}
At the implementation level, the frozen backbone together with the Task LoRA
serves as the shared demographic-independent reference around which the
Demographic LoRAs are composed. The Task LoRA captures the common survey-response protocol and valid output
format. Under the additional local-additivity approximation, the logit-space
derivation motivates the
main-effect parameterization
\begin{equation}
\Delta W_g(\mathbf v)
\approx\sum_{i=1}^{m}\Delta W_i(v_i).
\label{eq:additive_parameters}
\end{equation}
Equation~\ref{eq:additive_parameters} motivates the form of the composition;
it is not a claim that parameter addition is exactly equivalent to adding
complete-network logits.

In Stage~2, the active Demographic LoRAs are trained jointly on complete
respondent profiles. Consequently, an individual module should be interpreted
as a reusable component learned within an additive parameterization, rather
than as an independently identified or causal single-attribute effect. The
learned dimension-level weights further calibrate the relative contribution
of each demographic dimension.

The same construction makes an unseen demographic combination
constructible. Let $\mathbf v^{\mathrm{new}}$ be a complete profile absent
from training, and let $\mathcal M$ be the set of trained value-specific
modules. If every constituent value $v_i^{\mathrm{new}}$ occurred during
training, then $\Delta W_i(v_i^{\mathrm{new}})\in\mathcal M$ for every $i$.
The model can therefore form
\begin{equation}
\Delta W_{g^{\mathrm{new}}}
=\sum_{i=1}^{m}w_{d_i}\Delta W_i(v_i^{\mathrm{new}}),
\label{eq:unseen_composition}
\end{equation}
using the same learned dimension weights $w_{d_i}$. No complete-profile
adapter is required. This enables compositional construction of unseen
profiles from observed attribute values, with performance depending on module
coverage and on how well the additive approximation represents the relevant
demographic interactions.

\subsection{When Direct Addition Is Appropriate}

The parameter-space geometry of LoRA updates provides one way to interpret
possible redundancy or cancellation during composition. For two nonzero
updates, define their Frobenius cosine similarity as
\begin{equation}
\gamma_{i,j}=
\frac{\langle\Delta W_i,\Delta W_j\rangle_F}
{\|\Delta W_i\|_F\|\Delta W_j\|_F}.
\label{eq:gamma}
\end{equation}
Their combined magnitude follows from the inner-product expansion:
\begin{align}
\|\Delta W_i+\Delta W_j\|_F^2
&=\langle\Delta W_i+\Delta W_j,
\Delta W_i+\Delta W_j\rangle_F\\
&=\|\Delta W_i\|_F^2+\|\Delta W_j\|_F^2\notag\\
&\quad+2\langle\Delta W_i,\Delta W_j\rangle_F.
\label{eq:frob_sum}
\end{align}
Table~\ref{tab:geometry} summarizes several interpretive cases. Parameter
geometry alone does not determine complete-network behavior, but it can
indicate when direct addition may be redundant or partially cancelling.

\begin{table*}[t]
\small
\centering
\begin{tabular}{@{}L{0.15\textwidth}L{0.18\textwidth}L{0.39\textwidth}L{0.17\textwidth}@{}}
\toprule
Relation & Condition & Consequence & Possible treatment \\
\midrule
Approximately orthogonal & $\gamma_{i,j}\approx0$ & The updates have limited direct parameter-space overlap. & Direct addition \\
Same-direction redundancy & Large positive $\gamma_{i,j}$ & Similar shifts may be counted repeatedly and over-amplified. & Weighting or normalization \\
Opposite-direction conflict & Large negative $\gamma_{i,j}$ & The updates may partially cancel. & Weighting or orthogonal correction \\
Strong interaction & Additive main effects are insufficient & The joint occurrence may produce an effect not represented by the reusable components. & Interaction-aware extension \\
\bottomrule
\end{tabular}
\caption{Interpretive cases for LoRA composition. These parameter-space
relations do not by themselves guarantee the corresponding network-level
effects.}
\label{tab:geometry}
\end{table*}

These cases help delimit the current design. Let $\Delta W_{\ell,t}$ denote
the Task LoRA at layer $\ell$. It is trained on demographic-free survey
examples with balanced random valid option identifiers to learn the common
survey response protocol and valid output format, and is then frozen. For
each demographic value $(d,v)$, a Demographic LoRA contributes a reusable
component to the demographic-conditioned response pattern. Given profile
$g$, the effective layer weight is
\begin{equation}
\widetilde W_\ell(g)=W_\ell+\Delta W_{\ell,t}
+\sum_{(d,v)\in g}w_d\Delta W_{\ell,d,v},
\label{eq:parade_composition_definition}
\end{equation}
where $w_d=1+0.5\tanh(\eta_d)\in(0.5,1.5)$ is a learned weight shared by all
values of dimension $d$. To discourage a demographic module from reusing the
input subspace already represented by the Task LoRA, Stage~2 uses
\begin{equation}
\mathcal L_{\mathrm{orth}}(g)=
\frac{1}{|\mathcal L_A|\,|g|}
\sum_{\ell\in\mathcal L_A}\sum_{(d,v)\in g}
\frac{\|A_{\ell,t}A_{\ell,d,v}^{\top}\|_F^2}{r^2},
\label{eq:orthogonality_definition}
\end{equation}
where $\mathcal L_A$ is the set of adapted layers and $r$ is the LoRA rank.
The Stage~2 loss is the response loss plus
$\lambda\mathcal L_{\mathrm{orth}}$. The resulting model uses weighted
main-effect composition without explicit interaction modules. Dimension-level
weights can calibrate relative scales, but non-additive demographic
interactions remain outside the explicit parameterization.

\section{Implementation Details}
\label{app:implementation}

We report the dataset mappings, data splits, evaluation metrics, baseline
protocols, and optimization settings used in the experiments. The eight
demographic dimensions are selected based on their mutual information with
responses and their coverage across both datasets.

\subsection{Datasets and Demographic Schema}
\label{app:datasets}

\paragraph{World Values Survey.}
We use the WVS Wave~7 cross-national CSV, Datafile Version~6.0.0
(\texttt{WVS\_Cross-National\_Wave\_7\_csv\_v6\_0.csv}). The file contains 97,220 respondents. The
fixed question split contains 258 response questions: 207 for
training and 51 for testing. Administrative or demographic columns
(\texttt{Q260}--\texttt{Q290}), \texttt{Q223}, and \texttt{SDG1} are not
response targets. Blank values and negative numeric survey codes are treated
as missing; only answers present in the question's legal option map are
retained.

\paragraph{SocioBench.}
We use the 5,000-respondent, 555-question SocioBench ISSP subset, comprising 500 respondents in each of ten
domains: citizenship, environment, family, health, national identity,
religion, role of government, social inequality, social networks, and work
orientations. The fixed question split contains 441 training and 114 test questions.
The respective train/test question counts by domain are
47/12, 40/10, 43/11, 40/11, 48/12, 47/12, 48/12, 36/10, 47/12, and 45/12.
Country-specific option overrides are applied before validity checks.
Nonresponse codes and answers outside the applicable option map are excluded.
Special nonordinal codes are excluded from the distributional metrics but are
retained in MAE: an exact match receives zero error and a mismatch receives
unit error.

\paragraph{Demographic schema.}

We retain respondents with observed values for all eight dimensions in
training and evaluation. Table~\ref{tab:wvs_demographic_mapping} gives the WVS
mapping. A value is missing when its source field is blank, unparsable, or
uses a negative code. The suffixes shown in the table are the identifiers
used to name Demographic LoRAs.

\begin{table*}[t]
\small
\centering
\setlength{\tabcolsep}{3.5pt}
\begin{tabular}{@{}L{0.12\textwidth}L{0.15\textwidth}C{0.055\textwidth}L{0.595\textwidth}@{}}
\toprule
Dimension & WVS field & Count & Adapter values and mapping \\
\midrule
Gender & \texttt{Q260} & 2 &
\texttt{male}, \texttt{female}. \\
Age group & \texttt{Q262} & 3 &
\texttt{age\_young} ($\leq30$), \texttt{age\_middle} (31--55),
\texttt{age\_senior} ($\geq56$). \\
Country & \shortstack[l]{\texttt{B\_COUNTRY\_}\\\texttt{ALPHA}} & 66 &
\texttt{AND, ARG, ARM, AUS, BGD, BOL, BRA, CAN, CHL, CHN, COL, CYP, CZE,
DEU, ECU, EGY, ETH, GBR, GRC, GTM, HKG, IDN, IND, IRN, IRQ, JOR, JPN,
KAZ, KEN, KGZ, KOR, LBN, LBY, MAC, MAR, MDV, MEX, MMR, MNG, MYS, NGA,
NIC, NIR, NLD, NZL, PAK, PER, PHL, PRI, ROU, RUS, SGP, SRB, SVK, THA,
TJK, TUN, TUR, TWN, UKR, URY, USA, UZB, VEN, VNM, ZWE}. \\
Religion & \texttt{Q289} & 10 &
\texttt{rel\_0}--\texttt{rel\_9}: none, Catholic, Protestant, Orthodox,
Jewish, Muslim, Hindu, Buddhist, and the two source ``other'' codes. \\
Education & \texttt{Q275} & 9 &
\texttt{edu\_0}--\texttt{edu\_8}: early childhood/no education, primary,
lower secondary, upper secondary, post-secondary non-tertiary, short-cycle
tertiary, bachelor's, master's, and doctorate. \\
Marital status & \texttt{Q273} & 6 &
\texttt{marital\_married}, \texttt{marital\_living\_together},
\texttt{marital\_divorced}, \texttt{marital\_separated},
\texttt{marital\_widowed}, \texttt{marital\_single}. \\
Employment status & \texttt{Q279} & 8 &
\texttt{employment\_full\_time}, \texttt{employment\_part\_time},
\texttt{employment\_self\_employed}, \texttt{employment\_retired},
\texttt{employment\_homemaker}, \texttt{employment\_student},
\texttt{employment\_unemployed}, \texttt{employment\_other}. \\
Residence & \texttt{H\_URBRURAL} & 2 &
\texttt{urban}, \texttt{rural}. \\
\bottomrule
\end{tabular}
\caption{WVS demographic fields and the 106 observed attribute values used
to instantiate Demographic LoRAs.}
\label{tab:wvs_demographic_mapping}
\end{table*}

SocioBench is mapped into the same semantic schema. Free-text or
dataset-specific fields are normalized using the rules in
Table~\ref{tab:socio_demographic_mapping}. Education first uses reported
years of schooling when available. The intervals for
\texttt{edu\_0}--\texttt{edu\_8} are, respectively, $y\leq0$,
$0<y\leq6$, $6<y\leq9$, $9<y\leq12$, $12<y\leq14$, $14<y\leq16$,
$16<y\leq18$, $18<y\leq20$, and $y>20$ years; when years are unavailable,
the mapper uses the international education category. Paid work is full-time
at 30 or more reported hours per week and part-time below 30.

\begin{table*}[t]
\small
\centering
\setlength{\tabcolsep}{3.5pt}
\begin{tabular}{@{}L{0.12\textwidth}L{0.22\textwidth}C{0.055\textwidth}L{0.495\textwidth}@{}}
\toprule
Dimension & SocioBench field(s) & Count & Values observed in the fixed training split \\
\midrule
Gender & Sex of respondent & 2 & \texttt{male}, \texttt{female}. \\
Age group & Age of respondent & 3 &
\texttt{age\_young}, \texttt{age\_middle}, \texttt{age\_senior}, using the
same WVS cut points. \\
Country & Domain-specific country or survey-region field & 54 &
\texttt{AR, AT, AU, BE, BE\_BRU, BE\_FLA, BE\_WAL, BG, CA, CH, CL, CN,
CZ, DE, DE\_E, DE\_W, DK, EE, ES, FI, FR, GB\_GBN, GE, HR, HU, IE, IL,
IL\_A, IL\_J, IN, IS, IT, JP, KR, LT, LV, MX, NL, NO, NZ, PH, PL, PT,
RU, SE, SI, SK, SR, TH, TR, TW, US, VE, ZA}. \\
Religion & Affiliation/denomination & 9 &
\texttt{rel\_0}--\texttt{rel\_8}: none, Catholic, Protestant/Anglican,
Orthodox, Jewish, Muslim, Hindu, Buddhist, other religious affiliation. \\
Education & Years/category of education & 9 &
\texttt{edu\_0}--\texttt{edu\_8}, with the WVS-aligned labels. \\
Marital status & Legal partnership/status & 5 &
\texttt{marital\_married}, \texttt{marital\_divorced},
\texttt{marital\_separated}, \texttt{marital\_widowed},
\texttt{marital\_single}. \\
Employment status & Main status, paid work, hours & 6 &
\texttt{employment\_full\_time}, \texttt{employment\_part\_time},
\texttt{employment\_retired}, \texttt{employment\_student},
\texttt{employment\_unemployed}, \texttt{employment\_other}. \\
Residence & Urban--rural place of living & 2 &
\texttt{urban}, \texttt{rural}. \\
\bottomrule
\end{tabular}
\caption{SocioBench mapping and the 90 attribute values observed in the
fixed main-experiment training split. Country values may denote a country,
territory, or survey-region identifier. The mapper also recognizes
\texttt{marital\_living\_together}, \texttt{employment\_self\_employed},
and \texttt{employment\_homemaker}; these values did not instantiate
adapters because they were absent after the fixed split and completeness
filter.}
\label{tab:socio_demographic_mapping}
\end{table*}

WVS therefore trains
$2+3+66+10+9+6+8+2=106$ Demographic LoRAs and one shared Task LoRA,
for 107 LoRA modules in total. The Cartesian upper bound induced by the WVS
value sets contains
\begin{equation}
2\times3\times66\times10\times9\times6\times8\times2
=3{,}421{,}440
\end{equation}
complete profiles. A profile-specific scheme that assigns one equally sized
LoRA to every possible complete profile would therefore require 3,421,440
LoRAs, compared with 107 modules for \method{}. The complete-profile
inventory is therefore approximately 31,976 times larger than the total
\method{} adapter inventory. Comparing only demographic adapters,
3,421,440 profile adapters versus 106 Demographic LoRAs gives a 32,278-fold
difference. These counts instantiate the product-versus-sum complexity of profile-specific
and attribute-level adaptation. They describe the Cartesian upper bound rather
than the number of profiles observed in the training sample. If a
profile-specific scheme were restricted to the 678 profiles observed in the
selected WVS training sample, it would still require about 6.3 times as many
adapters as the 107 modules used by \method{}, while providing no adapter for
unseen combinations. These comparisons concern adapter inventory and
training/storage scaling rather than backbone computation for one survey
response.

For SocioBench, the corresponding counts are 90 Demographic LoRAs plus one
Task LoRA, compared with a Cartesian upper bound of 1,574,640 complete-profile
adapters.

\subsection{Data Splits}

\begin{table*}[t]
\small
\centering
\setlength{\tabcolsep}{3.3pt}
\begin{tabular}{@{}L{0.105\textwidth}C{0.105\textwidth}C{0.115\textwidth}C{0.115\textwidth}C{0.12\textwidth}C{0.12\textwidth}C{0.14\textwidth}@{}}
\toprule
Dataset & Raw respondents & Train respondents & Validation respondents &
Test respondents per seed & Train/test questions & Stage~1/Stage~2 train records \\
\midrule
WVS & 97,220 & 700 & 100 & 100 & 207/51 & 12,000/137,680 \\
SocioBench & 5,000 & 1,600 & 500 & 1,000 & 441/114 & 24,000/47,264 \\
\bottomrule
\end{tabular}
\caption{Fixed data splits. SocioBench counts are aggregated over ten
domains (160/50/100 train/validation/test respondents per domain).}
\label{tab:split_counts}
\end{table*}

\paragraph{WVS standard split.}
A fixed seed-42 partition assigns 96,220 of the 97,220 rows to the training
pool and 1,000 to the held-out pool. Requiring all eight demographic fields
leaves 92,469 training candidates; 700 are sampled with seed 42, yielding
678 distinct complete profiles. The validation split contains 100 complete
respondents sampled with seed 31,415. Training, validation, and final-test
respondent IDs are pairwise disjoint. For each evaluation seed in
$\{42,43,44,45,46\}$, 100 complete
respondents are drawn from the held-out pool. The same trained checkpoint is
reused across all five respondent-resampling seeds.

\paragraph{SocioBench standard split.}
Each domain contributes 160 of its 500 respondents to training, 50 to
validation, and 100 to each standard test split. The three respondent sets
are pairwise disjoint within each domain. A record is created only when both
its respondent and question belong to the corresponding split. Training uses
all ten domains jointly; respondent identities remain tied to their source
domain.

\paragraph{No train--test overlap.}
For both datasets, the training and final-test respondent-ID sets are
disjoint, and the training and test question-ID sets are disjoint. Thus, no
response from a final-test respondent and no test question is used in either
training stage. Validation respondents are also disjoint from training and
final-test respondents. The split is performed on IDs before response records are
constructed, rather than by randomly splitting respondent--question rows.

\paragraph{Unseen combinations.}
For an eligible unseen respondent, the full eight-way demographic tuple must
be absent from the selected training profiles while each of its eight
constituent attribute values must occur in training. This criterion produces
915 eligible WVS respondents, from which 100 are selected for each test
seed. SocioBench first applies the criterion within each domain and then
selects ten respondents per domain, for 100 unseen-combination respondents
per seed. These tests specifically evaluate recombination of observed
components, not
extrapolation to an unseen attribute value.

\subsection{Evaluation Metrics}
\label{app:metrics}

\paragraph{Individual-level metrics.}
Accuracy measures exact agreement with the observed response, whereas
normalized MAE measures how
far an ordinal prediction lies from the observed answer and makes questions
with different numbers of options comparable. Let $\mathcal E$ be all evaluated records and $\mathcal V\subseteq\mathcal E$
those with a valid parsed prediction. Invalid generations count as incorrect
for accuracy:
\begin{equation}
\operatorname{Acc}=\frac{1}{|\mathcal E|}
\sum_{i\in\mathcal E}\mathbf 1[\widehat y_i=y_i].
\end{equation}
MAE is averaged over valid predictions:
\begin{equation}
\operatorname{MAE}=\frac{1}{|\mathcal V|}\sum_{i\in\mathcal V}
\frac{|\operatorname{pos}_i(\widehat y_i)-
\operatorname{pos}_i(y_i)|}{K_i-1}.
\end{equation}
For WVS, \(\operatorname{pos}_i\) is the numeric option identifier. For
SocioBench it is the index in the contiguous ordinal core; a mismatch
involving a special nonordinal code receives unit error and such codes are
not included in the distribution metrics.

\paragraph{Population-level metrics.}
EMD measures how far probability mass must move along the ordered response
scale; JS divergence gives a bounded, symmetric measure of distributional
separation; and directional KL divergence penalizes simulated distributions
that assign too little mass to responses observed in the survey. The code
forms empirical and simulated distributions from frequencies of parsed
discrete predictions, not from averaged model probabilities. Cells are keyed by
dataset domain and question: WVS has one domain, while SocioBench retains its
ten domains. Thus these are question-level population distributions over the
sampled respondents, not separate complete-profile--question cells. Invalid
predictions are excluded. Every nonempty valid cell is retained, no minimum
cell-size threshold is imposed, and the final score is an unweighted macro
average over cells.

Complete eight-dimensional profiles are too sparse to estimate stable
empirical response distributions for most profile--question pairs. We
therefore evaluate the marginal population distribution for each question on
the sampled respondents. Individual-level Accuracy and MAE, together with the
unseen-profile evaluation, provide complementary evidence about demographic
conditioning and compositional generalization.

For a cell with $K$ ordered options, let $p$ and $\widehat p$ be the normalized
observed and predicted counts. Jensen--Shannon divergence uses base-2
logarithms:
\begin{equation}
\begin{aligned}
\operatorname{JSD}(p,\widehat p)
&=\tfrac12\operatorname{KL}_2(p\|m)
+\tfrac12\operatorname{KL}_2(\widehat p\|m),\\
m&=\tfrac12(p+\widehat p).
\end{aligned}
\end{equation}
For the reported directional KL, only the predicted distribution is floored:
\begin{equation}
\widehat p_k^{(\epsilon)}
=\frac{\max(\widehat p_k,\epsilon)}
{\sum_j\max(\widehat p_j,\epsilon)},\qquad \epsilon=10^{-8},
\end{equation}
\begin{equation}
\operatorname{KL}_2(p\|\widehat p^{(\epsilon)})
=\sum_{k:p_k>0}p_k\log_2
\frac{p_k}{\widehat p_k^{(\epsilon)}}.
\end{equation}
The normalized one-dimensional earth mover's distance is
\begin{equation}
\operatorname{EMD}(p,\widehat p)=
\frac{1}{K-1}\sum_{k=1}^{K-1}
\left|\sum_{j=1}^{k}p_j-\sum_{j=1}^{k}\widehat p_j\right|.
\end{equation}

Because our objective is population-level simulation, EMD, JS, and KL are
the primary metrics; Accuracy and MAE are retained to verify that improved
distribution matching does not come at the cost of poor individual-level
agreement.

\subsection{Baseline Specifications}

All methods use the same backbones, fixed respondent and question partitions,
survey records when training is required, candidate answer sets,
option-identifier response format, and evaluation protocol. All adapter-based
methods use LoRA rank $r=4$. For methods that use demographic conditioning,
the same eight demographic dimensions and normalized values are provided.
Method-specific conditioning and routing follow the original implementations.
To keep respondent-specific information consistent across methods, SimVBG is
not supplied with additional interaction history, since such history is not
provided to the other methods.

\subsection{Training and Inference Configuration}

In Stage~1, we sample
survey items from the training split, remove respondent demographics, and
replace each response with a valid option identifier sampled uniformly from
that question's candidate set. The resulting identifier frequencies are then
balanced across the constructed dataset to avoid encoding demographic
preference signals and systematic option-position bias. The constructed pool
is truncated to 12,000 WVS or 24,000 SocioBench records. Thus, sampling is
performed within the applicable candidate set of each question, while the
subsequent balancing is applied to identifier buckets over the constructed
Stage~1 dataset.

Stage~2 freezes the backbone and Task LoRA. Records are grouped by complete
profile so that each batch shares one active module composition; each survey
response remains a separate cross-entropy example. Only the eight active
Demographic LoRAs and their dimension-level weights are updated under
\begin{equation}
\mathcal L=\mathcal L_{\mathrm{CE}}+\lambda\mathcal L_{\mathrm{orth}}.
\end{equation}
At inference, the Task LoRA and the eight Demographic LoRAs selected by the
profile are composed once for that profile. The input contains only the survey
question and candidate answers, and the model outputs the selected option
identifier.

\paragraph{Prompt templates.}
Except for Persona Prompting and SimVBG, the remaining baselines and
\method{} use the same direct question--option prompt:
\begin{quote}
\small\ttfamily\raggedright
Question: \{question\}\\
Options:\\
\{option identifier\}. \{option text\}\\
\ldots\\[2pt]
Write ONLY your chosen option number.\\
Answer:
\end{quote}
Persona Prompting prepends the following structured profile:
\begin{quote}
\small\ttfamily\raggedright
You are a person with the following profile:\\
Gender: \{\}; Age: \{\}; Country: \{\}; Religion: \{\};\\
Education: \{\}; Marital status: \{\};\\
Employment status: \{\}; Place of living: \{\}.\\
You are a helpful assistant that answers survey questions honestly.
\end{quote}
SimVBG uses the same demographic information to generate a backstory and then
follows its original cognitive, affective, and behavioral prompting pipeline,
with a coordinator selecting the final option.

Experiments use Meta-Llama-3.1-8B-Instruct, Qwen3.5-9B, and the
Ministral-3-8B-Instruct-2512-BF16 checkpoint, with their corresponding
tokenizers.

Table~\ref{tab:hyperparameters} lists the primary \method{} hyperparameters
used across the six dataset--backbone settings unless otherwise specified.

\begin{table*}[t]
\small
\centering
\begin{tabular}{@{}L{0.29\textwidth}L{0.64\textwidth}@{}}
\toprule
\method{} setting & Value \\
\midrule
Hardware & Four NVIDIA A800-SXM4 GPUs with 80 GB memory per GPU. \\
LoRA configuration & Rank \(r=4\), \(\alpha=8\), dropout 0.05; Kaiming-uniform
initialization for \(A\) and zero initialization for \(B\). \\
Target projections & Attention \texttt{q\_proj}, \texttt{k\_proj}, and
\texttt{v\_proj}; MLP \texttt{gate\_proj},
\texttt{up\_proj}, \texttt{down\_proj}. \\
Optimizer & AdamW; learning rate \(10^{-4}\) for LoRA factors and dimension
weights; weight decay 0.01. \\
Schedule & Cosine decay with a 3\% warmup fraction. \\
Training epochs & One epoch in Stage~1 and one epoch in Stage~2. \\
Maximum sequence length & 1,024 tokens. \\
Precision & BFloat16. \\
Gradient clipping & Global norm 1.0. \\
Supervision & Cross-entropy loss only on target option-identifier tokens;
prompt tokens are masked. \\
Dimension weights & One scalar per dimension,
\(w_d=1+0.5\tanh(\eta_d)\in(0.5,1.5)\), initialized with \(\eta_d=0\). \\
Randomness & Training and split seed 42; final test-respondent resampling seeds
42--46; one fixed trained checkpoint per setting. \\
\bottomrule
\end{tabular}
\caption{Primary \method{} hyperparameters used across datasets and
backbones unless otherwise specified.}
\label{tab:hyperparameters}
\end{table*}

The effective global batch size is 64 for WVS and 32 for SocioBench.
SocioBench Ministral Stage~2 uses gradient accumulation to obtain an effective
batch size of 64. Each stage makes one pass over the record counts in
Table~\ref{tab:split_counts}.

The implemented layer update is
\begin{equation}
\widetilde W_\ell(g)=W_\ell+s_{\mathrm{task}}\Delta W_{\ell,t}
+s_{\mathrm{demo}}\sum_{(d,v)\in g}w_d\Delta W_{\ell,d,v},
\end{equation}
where \(s_{\mathrm{task}}\) and \(s_{\mathrm{demo}}\) are fixed global
scales. They are distinct from the learned dimension weights \(w_d\).
Equivalently, these fixed scales can be absorbed into the definitions of the
corresponding LoRA updates.

\begin{table*}[t]
\small
\centering
\begin{tabular}{@{}L{0.17\textwidth}L{0.29\textwidth}C{0.12\textwidth}C{0.14\textwidth}C{0.18\textwidth}@{}}
\toprule
Dataset & Backbone & \(\lambda\) & Task scale & Demographic scale \\
\midrule
WVS & Llama-3.1-8B-Instruct & 0.3 & 0.5 & 1.0 \\
WVS & Qwen3.5-9B & 0.5 & 1.0 & 0.75 \\
WVS & Ministral-3-8B-Instruct & 0.3 & 1.0 & 1.1 \\
SocioBench & Llama-3.1-8B-Instruct & 0.4 & 0.75 & 0.66 \\
SocioBench & Qwen3.5-9B & 0.7 & 0.75 & 1.15 \\
SocioBench & Ministral-3-8B-Instruct & 0.5 & 0.875 & 1.0 \\
\bottomrule
\end{tabular}
\caption{Dataset--backbone-specific orthogonality coefficients and global
composition scales used in the reported experiments.}
\label{tab:final_lambda_scales}
\end{table*}

\section{Detailed Experimental Results}
\label{app:detailed_results}

The first two subsections report results for all five test-respondent
resampling seeds, 42--46. Only the evaluation
respondents are resampled: each trainable method uses one fixed checkpoint,
and the test-question set remains unchanged across seeds. Bold and underlined entries
are recomputed within each backbone--dataset--metric block, and \method{} is
shaded in yellow.

\subsection{Standard Test Split}

Tables~\ref{tab:standard_seed_42_results}--\ref{tab:standard_seed_46_results}
report the per-seed values for the five evaluation seeds used in the main
results table. Within a
seed, every method is evaluated on the same respondent and question
splits. Across seeds, the trained parameters and question set stay fixed;
only the held-out respondent sample changes.

\begin{table*}[t]
\small
\centering
\setlength{\tabcolsep}{3.2pt}
\resizebox{\textwidth}{!}{%
\begin{tabular}{@{}llcccccccccc@{}}
\toprule
\textbf{Backbone}
& \textbf{Method}
& \multicolumn{5}{c}{\textbf{WVS}}
& \multicolumn{5}{c}{\textbf{SocioBench}} \\
\cmidrule(lr){3-7}\cmidrule(lr){8-12}
&
& \textbf{Acc. (\%) $\uparrow$}
& \textbf{MAE $\downarrow$}
& \textbf{EMD $\downarrow$}
& \textbf{JS $\downarrow$}
& \textbf{KL $\downarrow$}
& \textbf{Acc. (\%) $\uparrow$}
& \textbf{MAE $\downarrow$}
& \textbf{EMD $\downarrow$}
& \textbf{JS $\downarrow$}
& \textbf{KL $\downarrow$} \\
\midrule
\multirow{7}{*}{\shortstack{Llama-3.1-8B-\\Instruct}} & Vanilla Baseline & 37.55 & 0.3315 & 0.3312 & 0.4595 & 14.8718 & 31.94 & 0.2959 & 0.2897 & 0.4926 & 16.1477 \\

 & Persona Prompting & 36.73 & 0.3204 & 0.2710 & 0.3708 & 10.5493 & 34.36 & 0.2652 & \underline{0.2162} & 0.3227 & 8.8838 \\

 & SimVBG & 41.96 & 0.3286 & 0.2706 & 0.2899 & 7.4783 & 31.45 & 0.3144 & 0.2397 & \underline{0.2894} & \textbf{5.5617} \\

 & LoRA & 47.88 & \underline{0.2958} & 0.2872 & 0.3317 & 11.1540 & 37.01 & 0.2690 & 0.2615 & 0.4340 & 14.7596 \\

 & MixLoRA & \textbf{48.43} & \underline{0.2958} & 0.2829 & 0.3137 & 10.5168 & \textbf{38.18} & \textbf{0.2594} & 0.2492 & 0.4076 & 13.8942 \\

 & CuMA & 46.65 & 0.3079 & \underline{0.2458} & \underline{0.2505} & \underline{7.3939} & 37.78 & 0.2652 & 0.2187 & 0.3217 & 9.2479 \\

\rowcolor{paradeyellow}
\cellcolor{white}
 & \textbf{PARADE} & \underline{48.10} & \textbf{0.2838} & \textbf{0.1755} & \textbf{0.1802} & \textbf{4.9530} & \underline{38.15} & \underline{0.2632} & \textbf{0.1877} & \textbf{0.2411} & \underline{6.0420} \\

\midrule

\multirow{7}{*}{Qwen3.5-9B} & Vanilla Baseline & 40.59 & 0.3458 & 0.3455 & 0.4427 & 14.1064 & 32.74 & 0.2977 & 0.2905 & 0.4848 & 15.9049 \\

 & Persona Prompting & 40.29 & 0.3238 & 0.2757 & 0.3425 & 8.2352 & 33.96 & 0.2711 & 0.2229 & 0.3529 & 9.8498 \\

 & SimVBG & 42.49 & 0.3060 & \underline{0.2177} & \underline{0.2314} & \underline{6.0659} & 35.54 & 0.2643 & \underline{0.1821} & \underline{0.2337} & \underline{4.6381} \\

 & LoRA & 45.78 & 0.3235 & 0.3226 & 0.3720 & 12.5233 & \underline{37.38} & 0.2667 & 0.2529 & 0.4266 & 14.4926 \\

 & MixLoRA & 46.94 & 0.3131 & 0.3078 & 0.3534 & 11.8441 & 36.95 & 0.2755 & 0.2562 & 0.4177 & 14.2469 \\

 & CuMA & \underline{47.53} & \underline{0.2988} & 0.2317 & 0.2490 & 6.5341 & 37.25 & \underline{0.2617} & 0.2183 & 0.3160 & 8.9844 \\

\rowcolor{paradeyellow}
\cellcolor{white}
 & \textbf{PARADE} & \textbf{48.20} & \textbf{0.2916} & \textbf{0.1734} & \textbf{0.1947} & \textbf{5.1220} & \textbf{38.79} & \textbf{0.2526} & \textbf{0.1647} & \textbf{0.1946} & \textbf{4.3586} \\

\midrule

\multirow{7}{*}{\shortstack{Ministral-3-8B-\\Instruct}} & Vanilla Baseline & 36.61 & 0.3332 & 0.3328 & 0.4815 & 15.1319 & 32.18 & 0.2815 & 0.2675 & 0.4836 & 15.9761 \\

 & Persona Prompting & 38.31 & 0.3028 & 0.2299 & 0.2991 & 7.5389 & 33.67 & 0.2641 & 0.2082 & 0.3404 & 9.7428 \\

 & SimVBG & 37.53 & 0.3210 & 0.2215 & 0.2362 & \underline{4.7625} & 33.96 & 0.2785 & \textbf{0.1639} & \textbf{0.1739} & \textbf{3.1069} \\

 & LoRA & 47.35 & 0.3095 & 0.3065 & 0.3543 & 12.0765 & 37.38 & 0.2714 & 0.2687 & 0.4328 & 14.6738 \\

 & MixLoRA & \underline{47.80} & 0.3051 & 0.2782 & 0.3256 & 11.0253 & 36.78 & 0.2782 & 0.2591 & 0.4217 & 14.2452 \\

 & CuMA & \textbf{48.84} & \underline{0.3015} & \underline{0.2065} & \underline{0.2056} & 5.6516 & \textbf{38.85} & \textbf{0.2558} & 0.2080 & 0.2866 & 7.8963 \\

\rowcolor{paradeyellow}
\cellcolor{white}
 & \textbf{PARADE} & 47.00 & \textbf{0.2950} & \textbf{0.1356} & \textbf{0.1122} & \textbf{2.2750} & \underline{37.81} & \underline{0.2597} & \underline{0.1732} & \underline{0.2055} & \underline{4.1909} \\

\bottomrule
\end{tabular}%
}
\caption{Results for test-respondent resampling seed 42 on the standard WVS and SocioBench test splits. Bold and underlined values denote the best and second-best results within each backbone and dataset, respectively. The \method{} rows are highlighted in yellow.}
\label{tab:standard_seed_42_results}
\end{table*}

\begin{table*}[t]
\small
\centering
\setlength{\tabcolsep}{3.2pt}
\resizebox{\textwidth}{!}{%
\begin{tabular}{@{}llcccccccccc@{}}
\toprule
\textbf{Backbone}
& \textbf{Method}
& \multicolumn{5}{c}{\textbf{WVS}}
& \multicolumn{5}{c}{\textbf{SocioBench}} \\
\cmidrule(lr){3-7}\cmidrule(lr){8-12}
&
& \textbf{Acc. (\%) $\uparrow$}
& \textbf{MAE $\downarrow$}
& \textbf{EMD $\downarrow$}
& \textbf{JS $\downarrow$}
& \textbf{KL $\downarrow$}
& \textbf{Acc. (\%) $\uparrow$}
& \textbf{MAE $\downarrow$}
& \textbf{EMD $\downarrow$}
& \textbf{JS $\downarrow$}
& \textbf{KL $\downarrow$} \\
\midrule
\multirow{7}{*}{\shortstack{Llama-3.1-8B-\\Instruct}} & Vanilla Baseline & 38.51 & 0.3294 & 0.3287 & 0.4538 & 14.6621 & 31.95 & 0.2977 & 0.2914 & 0.4920 & 16.1400 \\

 & Persona Prompting & 37.20 & 0.3210 & 0.2695 & 0.3621 & 9.1599 & 34.67 & 0.2662 & \underline{0.2161} & 0.3182 & 8.6742 \\

 & SimVBG & 42.32 & 0.3353 & 0.2748 & 0.2803 & 7.3518 & 31.45 & 0.3131 & 0.2384 & \underline{0.2861} & \textbf{5.5659} \\

 & LoRA & \textbf{49.20} & \underline{0.2965} & 0.2881 & 0.3276 & 10.9649 & 37.50 & 0.2700 & 0.2630 & 0.4301 & 14.6408 \\

 & MixLoRA & \underline{48.98} & 0.2999 & 0.2846 & 0.3100 & 10.4311 & \textbf{38.48} & \textbf{0.2624} & 0.2519 & 0.4054 & 13.8278 \\

 & CuMA & 47.03 & 0.3140 & \underline{0.2534} & \underline{0.2442} & \underline{6.9689} & \underline{38.22} & 0.2657 & 0.2197 & 0.3196 & 9.1003 \\

\rowcolor{paradeyellow}
\cellcolor{white}
 & \textbf{PARADE} & 48.65 & \textbf{0.2859} & \textbf{0.1734} & \textbf{0.1766} & \textbf{4.6757} & 38.06 & \underline{0.2655} & \textbf{0.1896} & \textbf{0.2427} & \underline{6.0880} \\

\midrule

\multirow{7}{*}{Qwen3.5-9B} & Vanilla Baseline & 40.09 & 0.3460 & 0.3457 & 0.4510 & 14.0834 & 32.04 & 0.3020 & 0.2954 & 0.4934 & 16.1037 \\

 & Persona Prompting & 40.36 & 0.3259 & 0.2749 & 0.3386 & 8.2555 & 33.98 & 0.2723 & 0.2262 & 0.3602 & 10.1852 \\

 & SimVBG & 44.45 & 0.3009 & \underline{0.2179} & \underline{0.2210} & \underline{5.8985} & 35.85 & 0.2636 & \underline{0.1822} & \underline{0.2327} & \underline{4.4361} \\

 & LoRA & 47.58 & 0.3202 & 0.3184 & 0.3586 & 12.0963 & 37.63 & 0.2692 & 0.2559 & 0.4253 & 14.4365 \\

 & MixLoRA & 48.53 & 0.3084 & 0.3037 & 0.3417 & 11.4171 & 36.55 & 0.2803 & 0.2616 & 0.4231 & 14.3440 \\

 & CuMA & \underline{49.00} & \underline{0.2957} & 0.2281 & 0.2379 & 6.6567 & \underline{37.78} & \underline{0.2635} & 0.2219 & 0.3149 & 8.5833 \\

\rowcolor{paradeyellow}
\cellcolor{white}
 & \textbf{PARADE} & \textbf{49.73} & \textbf{0.2884} & \textbf{0.1816} & \textbf{0.1909} & \textbf{4.8626} & \textbf{38.82} & \textbf{0.2541} & \textbf{0.1670} & \textbf{0.1943} & \textbf{4.2305} \\

\midrule

\multirow{7}{*}{\shortstack{Ministral-3-8B-\\Instruct}} & Vanilla Baseline & 37.26 & 0.3348 & 0.3351 & 0.4803 & 15.0505 & 32.38 & 0.2831 & 0.2695 & 0.4823 & 15.9319 \\

 & Persona Prompting & 38.82 & 0.3048 & 0.2406 & 0.3026 & 7.7188 & 33.79 & 0.2633 & 0.2084 & 0.3438 & 9.6672 \\

 & SimVBG & 38.36 & 0.3224 & 0.2251 & 0.2290 & \underline{4.5538} & 34.29 & 0.2754 & \textbf{0.1625} & \textbf{0.1710} & \textbf{2.6247} \\

 & LoRA & 48.73 & 0.3084 & 0.3060 & 0.3491 & 11.8201 & 37.57 & 0.2716 & 0.2697 & 0.4319 & 14.6533 \\

 & MixLoRA & \underline{49.47} & 0.3013 & 0.2771 & 0.3195 & 10.7800 & 36.75 & 0.2803 & 0.2616 & 0.4221 & 14.2444 \\

 & CuMA & \textbf{50.53} & \underline{0.2953} & \underline{0.2106} & \underline{0.2014} & 5.6304 & \textbf{38.72} & \textbf{0.2568} & 0.2073 & 0.2880 & 7.9377 \\

\rowcolor{paradeyellow}
\cellcolor{white}
 & \textbf{PARADE} & 48.08 & \textbf{0.2929} & \textbf{0.1261} & \textbf{0.1102} & \textbf{2.2036} & \underline{38.01} & \underline{0.2605} & \underline{0.1738} & \underline{0.2007} & \underline{4.0112} \\

\bottomrule
\end{tabular}%
}
\caption{Results for test-respondent resampling seed 43 on the standard WVS and SocioBench test splits. Bold and underlined values denote the best and second-best results within each backbone and dataset, respectively. The \method{} rows are highlighted in yellow.}
\label{tab:standard_seed_43_results}
\end{table*}

\begin{table*}[t]
\small
\centering
\setlength{\tabcolsep}{3.2pt}
\resizebox{\textwidth}{!}{%
\begin{tabular}{@{}llcccccccccc@{}}
\toprule
\textbf{Backbone}
& \textbf{Method}
& \multicolumn{5}{c}{\textbf{WVS}}
& \multicolumn{5}{c}{\textbf{SocioBench}} \\
\cmidrule(lr){3-7}\cmidrule(lr){8-12}
&
& \textbf{Acc. (\%) $\uparrow$}
& \textbf{MAE $\downarrow$}
& \textbf{EMD $\downarrow$}
& \textbf{JS $\downarrow$}
& \textbf{KL $\downarrow$}
& \textbf{Acc. (\%) $\uparrow$}
& \textbf{MAE $\downarrow$}
& \textbf{EMD $\downarrow$}
& \textbf{JS $\downarrow$}
& \textbf{KL $\downarrow$} \\
\midrule
\multirow{7}{*}{\shortstack{Llama-3.1-8B-\\Instruct}} & Vanilla Baseline & 36.54 & 0.3391 & 0.3382 & 0.4694 & 15.1326 & 32.10 & 0.2943 & 0.2881 & 0.4907 & 16.1079 \\

 & Persona Prompting & 36.43 & 0.3235 & 0.2707 & 0.3654 & 9.9210 & 34.54 & \underline{0.2649} & \underline{0.2169} & 0.3221 & 8.9748 \\

 & SimVBG & 40.32 & 0.3432 & 0.2830 & 0.2958 & 7.7038 & 31.03 & 0.3155 & 0.2379 & \underline{0.2864} & \textbf{5.7463} \\

 & LoRA & 46.86 & 0.3054 & 0.2956 & 0.3463 & 11.5264 & 37.09 & 0.2696 & 0.2628 & 0.4343 & 14.7607 \\

 & MixLoRA & \underline{47.13} & \underline{0.3046} & 0.2876 & 0.3222 & 10.7259 & \textbf{38.16} & \textbf{0.2628} & 0.2524 & 0.4101 & 13.9520 \\

 & CuMA & 45.69 & 0.3161 & \underline{0.2523} & \underline{0.2533} & \underline{7.3796} & \underline{37.82} & 0.2663 & 0.2200 & 0.3221 & 9.2720 \\

\rowcolor{paradeyellow}
\cellcolor{white}
 & \textbf{PARADE} & \textbf{47.60} & \textbf{0.2879} & \textbf{0.1756} & \textbf{0.1879} & \textbf{5.0966} & 37.75 & 0.2651 & \textbf{0.1905} & \textbf{0.2454} & \underline{6.1722} \\

\midrule

\multirow{7}{*}{Qwen3.5-9B} & Vanilla Baseline & 40.24 & 0.3443 & 0.3439 & 0.4429 & 13.9754 & 32.14 & 0.3011 & 0.2947 & 0.4926 & 16.0871 \\

 & Persona Prompting & 40.18 & 0.3221 & 0.2677 & 0.3308 & 8.5587 & 33.75 & 0.2720 & 0.2247 & 0.3587 & 10.3564 \\

 & SimVBG & 41.84 & 0.3135 & \underline{0.2269} & \underline{0.2299} & \underline{5.9890} & 35.41 & 0.2651 & \underline{0.1818} & \underline{0.2323} & \textbf{4.4379} \\

 & LoRA & 45.79 & 0.3225 & 0.3206 & 0.3736 & 12.5240 & \underline{37.54} & 0.2676 & 0.2546 & 0.4257 & 14.4783 \\

 & MixLoRA & \underline{46.70} & 0.3133 & 0.3067 & 0.3569 & 11.8672 & 36.91 & 0.2769 & 0.2587 & 0.4197 & 14.2919 \\

 & CuMA & 46.60 & \underline{0.3075} & 0.2409 & 0.2656 & 7.2321 & 37.46 & \underline{0.2638} & 0.2223 & 0.3192 & 8.9929 \\

\rowcolor{paradeyellow}
\cellcolor{white}
 & \textbf{PARADE} & \textbf{47.99} & \textbf{0.2928} & \textbf{0.1743} & \textbf{0.1999} & \textbf{5.4496} & \textbf{38.61} & \textbf{0.2524} & \textbf{0.1669} & \textbf{0.2015} & \underline{4.4773} \\

\midrule

\multirow{7}{*}{\shortstack{Ministral-3-8B-\\Instruct}} & Vanilla Baseline & 36.66 & 0.3368 & 0.3364 & 0.4795 & 15.1368 & 31.73 & 0.2844 & 0.2709 & 0.4894 & 16.1130 \\

 & Persona Prompting & 38.16 & 0.3109 & 0.2389 & 0.2964 & 7.4673 & 33.47 & 0.2646 & 0.2062 & 0.3384 & 9.5564 \\

 & SimVBG & 36.20 & 0.3337 & 0.2317 & 0.2440 & \underline{5.5708} & 34.27 & 0.2768 & \textbf{0.1610} & \textbf{0.1738} & \textbf{2.9734} \\

 & LoRA & 47.09 & 0.3106 & 0.3083 & 0.3612 & 12.2098 & 37.65 & 0.2709 & 0.2683 & 0.4298 & 14.6263 \\

 & MixLoRA & \underline{47.40} & 0.3082 & 0.2910 & 0.3413 & 11.3741 & 36.73 & 0.2776 & 0.2588 & 0.4219 & 14.2787 \\

 & CuMA & \textbf{48.34} & \textbf{0.3024} & \underline{0.2073} & \underline{0.2088} & 6.0273 & \textbf{38.64} & \textbf{0.2552} & 0.2057 & 0.2854 & 7.8617 \\

\rowcolor{paradeyellow}
\cellcolor{white}
 & \textbf{PARADE} & 46.47 & \underline{0.3040} & \textbf{0.1401} & \textbf{0.1256} & \textbf{2.5130} & \underline{38.02} & \underline{0.2587} & \underline{0.1710} & \underline{0.2013} & \underline{4.0789} \\

\bottomrule
\end{tabular}%
}
\caption{Results for test-respondent resampling seed 44 on the standard WVS and SocioBench test splits. Bold and underlined values denote the best and second-best results within each backbone and dataset, respectively. The \method{} rows are highlighted in yellow.}
\label{tab:standard_seed_44_results}
\end{table*}

\begin{table*}[t]
\small
\centering
\setlength{\tabcolsep}{3.2pt}
\resizebox{\textwidth}{!}{%
\begin{tabular}{@{}llcccccccccc@{}}
\toprule
\textbf{Backbone}
& \textbf{Method}
& \multicolumn{5}{c}{\textbf{WVS}}
& \multicolumn{5}{c}{\textbf{SocioBench}} \\
\cmidrule(lr){3-7}\cmidrule(lr){8-12}
&
& \textbf{Acc. (\%) $\uparrow$}
& \textbf{MAE $\downarrow$}
& \textbf{EMD $\downarrow$}
& \textbf{JS $\downarrow$}
& \textbf{KL $\downarrow$}
& \textbf{Acc. (\%) $\uparrow$}
& \textbf{MAE $\downarrow$}
& \textbf{EMD $\downarrow$}
& \textbf{JS $\downarrow$}
& \textbf{KL $\downarrow$} \\
\midrule
\multirow{7}{*}{\shortstack{Llama-3.1-8B-\\Instruct}} & Vanilla Baseline & 39.06 & 0.3244 & 0.3241 & 0.4476 & 14.4880 & 32.32 & 0.2940 & 0.2876 & 0.4883 & 16.0406 \\

 & Persona Prompting & 38.41 & 0.3093 & 0.2580 & 0.3560 & 9.3962 & 34.87 & 0.2651 & \underline{0.2140} & 0.3151 & 8.8519 \\

 & SimVBG & 42.28 & 0.3252 & 0.2608 & 0.2675 & \underline{6.6467} & 30.43 & 0.3179 & 0.2400 & \underline{0.2886} & \textbf{5.5615} \\

 & LoRA & 48.55 & \underline{0.2967} & 0.2858 & 0.3291 & 10.7374 & 37.57 & 0.2678 & 0.2601 & 0.4288 & 14.6054 \\

 & MixLoRA & \underline{48.78} & \underline{0.2967} & 0.2819 & 0.3120 & 10.4151 & \textbf{38.34} & \textbf{0.2616} & 0.2501 & 0.4068 & 13.8549 \\

 & CuMA & 46.83 & 0.3094 & \underline{0.2410} & \underline{0.2407} & 7.0346 & 38.11 & 0.2657 & 0.2175 & 0.3163 & 9.0907 \\

\rowcolor{paradeyellow}
\cellcolor{white}
 & \textbf{PARADE} & \textbf{49.75} & \textbf{0.2763} & \textbf{0.1677} & \textbf{0.1725} & \textbf{4.7463} & \underline{38.26} & \underline{0.2647} & \textbf{0.1869} & \textbf{0.2445} & \underline{6.2102} \\

\midrule

\multirow{7}{*}{Qwen3.5-9B} & Vanilla Baseline & 39.66 & 0.3473 & 0.3473 & 0.4554 & 14.1321 & 32.00 & 0.3017 & 0.2942 & 0.4931 & 16.0988 \\

 & Persona Prompting & 39.72 & 0.3277 & 0.2779 & 0.3439 & 7.9405 & 33.89 & 0.2718 & 0.2233 & 0.3557 & 10.1230 \\

 & SimVBG & 43.59 & 0.3038 & \underline{0.2262} & \underline{0.2286} & \underline{5.5140} & 35.60 & 0.2655 & \underline{0.1818} & \underline{0.2316} & \underline{4.7189} \\

 & LoRA & 48.27 & 0.3104 & 0.3091 & 0.3487 & 11.9038 & \underline{37.88} & 0.2667 & 0.2530 & 0.4221 & 14.3674 \\

 & MixLoRA & \underline{49.02} & 0.3011 & 0.2962 & 0.3335 & 11.2827 & 37.40 & 0.2756 & 0.2572 & 0.4153 & 14.1604 \\

 & CuMA & 48.39 & \underline{0.2964} & 0.2328 & 0.2409 & 6.1866 & 37.87 & \underline{0.2637} & 0.2206 & 0.3134 & 8.8907 \\

\rowcolor{paradeyellow}
\cellcolor{white}
 & \textbf{PARADE} & \textbf{49.48} & \textbf{0.2862} & \textbf{0.1722} & \textbf{0.1957} & \textbf{5.5050} & \textbf{39.20} & \textbf{0.2556} & \textbf{0.1679} & \textbf{0.1960} & \textbf{4.4756} \\

\midrule

\multirow{7}{*}{\shortstack{Ministral-3-8B-\\Instruct}} & Vanilla Baseline & 37.27 & 0.3297 & 0.3293 & 0.4793 & 14.9818 & 32.40 & 0.2820 & 0.2676 & 0.4808 & 15.9067 \\

 & Persona Prompting & 39.56 & 0.2982 & 0.2290 & 0.2939 & 7.1636 & 34.02 & 0.2631 & 0.2030 & 0.3369 & 9.5497 \\

 & SimVBG & 38.60 & 0.3200 & 0.2247 & 0.2306 & \underline{4.7825} & 34.20 & 0.2789 & \textbf{0.1632} & \textbf{0.1707} & \textbf{2.7370} \\

 & LoRA & 48.67 & 0.3039 & 0.3015 & 0.3449 & 11.7700 & 38.01 & 0.2700 & 0.2668 & 0.4264 & 14.5063 \\

 & MixLoRA & \textbf{49.50} & 0.2993 & 0.2797 & 0.3183 & 10.8218 & 37.20 & 0.2768 & 0.2578 & 0.4194 & 14.1737 \\

 & CuMA & \underline{49.44} & \underline{0.2964} & \underline{0.2083} & \underline{0.2003} & 5.4449 & \textbf{38.79} & \textbf{0.2563} & 0.2053 & 0.2829 & 7.7097 \\

\rowcolor{paradeyellow}
\cellcolor{white}
 & \textbf{PARADE} & 48.67 & \textbf{0.2893} & \textbf{0.1244} & \textbf{0.1116} & \textbf{2.3750} & \underline{38.38} & \underline{0.2606} & \underline{0.1697} & \underline{0.1957} & \underline{3.9777} \\

\bottomrule
\end{tabular}%
}
\caption{Results for test-respondent resampling seed 45 on the standard WVS and SocioBench test splits. Bold and underlined values denote the best and second-best results within each backbone and dataset, respectively. The \method{} rows are highlighted in yellow.}
\label{tab:standard_seed_45_results}
\end{table*}

\begin{table*}[t]
\small
\centering
\setlength{\tabcolsep}{3.2pt}
\resizebox{\textwidth}{!}{%
\begin{tabular}{@{}llcccccccccc@{}}
\toprule
\textbf{Backbone}
& \textbf{Method}
& \multicolumn{5}{c}{\textbf{WVS}}
& \multicolumn{5}{c}{\textbf{SocioBench}} \\
\cmidrule(lr){3-7}\cmidrule(lr){8-12}
&
& \textbf{Acc. (\%) $\uparrow$}
& \textbf{MAE $\downarrow$}
& \textbf{EMD $\downarrow$}
& \textbf{JS $\downarrow$}
& \textbf{KL $\downarrow$}
& \textbf{Acc. (\%) $\uparrow$}
& \textbf{MAE $\downarrow$}
& \textbf{EMD $\downarrow$}
& \textbf{JS $\downarrow$}
& \textbf{KL $\downarrow$} \\
\midrule
\multirow{7}{*}{\shortstack{Llama-3.1-8B-\\Instruct}} & Vanilla Baseline & 37.88 & 0.3266 & 0.3260 & 0.4585 & 14.8175 & 31.49 & 0.2982 & 0.2922 & 0.4970 & 16.2696 \\

 & Persona Prompting & 38.12 & 0.3108 & 0.2593 & 0.3487 & 9.0761 & 34.50 & 0.2661 & \underline{0.2158} & 0.3155 & 8.5096 \\

 & SimVBG & 42.49 & 0.3237 & 0.2631 & 0.2729 & 7.0828 & 31.15 & 0.3151 & 0.2367 & \underline{0.2787} & \textbf{5.1174} \\

 & LoRA & \underline{48.06} & \underline{0.2955} & 0.2857 & 0.3357 & 10.9288 & 37.20 & 0.2686 & 0.2610 & 0.4319 & 14.7024 \\

 & MixLoRA & \textbf{48.22} & 0.2982 & 0.2805 & 0.3163 & 10.2374 & \underline{37.94} & \textbf{0.2617} & 0.2505 & 0.4102 & 13.9632 \\

 & CuMA & 46.72 & 0.3080 & \underline{0.2481} & \underline{0.2455} & \underline{6.7541} & \textbf{38.33} & \underline{0.2640} & 0.2186 & 0.3199 & 9.1057 \\

\rowcolor{paradeyellow}
\cellcolor{white}
 & \textbf{PARADE} & 47.82 & \textbf{0.2848} & \textbf{0.1694} & \textbf{0.1859} & \textbf{5.2456} & 37.59 & 0.2659 & \textbf{0.1892} & \textbf{0.2447} & \underline{6.2114} \\

\midrule

\multirow{7}{*}{Qwen3.5-9B} & Vanilla Baseline & 40.46 & 0.3390 & 0.3384 & 0.4440 & 13.9529 & 32.10 & 0.3020 & 0.2947 & 0.4926 & 16.0712 \\

 & Persona Prompting & 41.13 & 0.3144 & 0.2633 & 0.3252 & 7.9659 & 33.44 & 0.2747 & 0.2244 & 0.3576 & 10.0790 \\

 & SimVBG & 44.40 & \underline{0.2946} & \underline{0.2096} & \underline{0.2112} & \underline{5.3532} & 35.20 & 0.2665 & \underline{0.1849} & \underline{0.2362} & \underline{4.7445} \\

 & LoRA & 46.41 & 0.3217 & 0.3203 & 0.3698 & 12.3655 & 37.42 & 0.2677 & 0.2537 & 0.4263 & 14.4766 \\

 & MixLoRA & 47.21 & 0.3144 & 0.3069 & 0.3513 & 11.7005 & 36.66 & 0.2778 & 0.2591 & 0.4216 & 14.3395 \\

 & CuMA & \underline{47.92} & 0.2980 & 0.2358 & 0.2503 & 7.0079 & \underline{37.45} & \underline{0.2632} & 0.2218 & 0.3186 & 8.8769 \\

\rowcolor{paradeyellow}
\cellcolor{white}
 & \textbf{PARADE} & \textbf{48.95} & \textbf{0.2868} & \textbf{0.1796} & \textbf{0.1956} & \textbf{5.0031} & \textbf{38.45} & \textbf{0.2541} & \textbf{0.1661} & \textbf{0.1959} & \textbf{4.3834} \\

\midrule

\multirow{7}{*}{\shortstack{Ministral-3-8B-\\Instruct}} & Vanilla Baseline & 37.23 & 0.3271 & 0.3266 & 0.4760 & 15.0154 & 32.08 & 0.2841 & 0.2704 & 0.4849 & 16.0058 \\

 & Persona Prompting & 39.85 & 0.2946 & 0.2279 & 0.2899 & 7.1953 & 33.62 & 0.2660 & 0.2070 & 0.3340 & 9.4973 \\

 & SimVBG & 38.24 & 0.3194 & 0.2191 & 0.2298 & \underline{5.1258} & 34.09 & 0.2795 & \textbf{0.1638} & \textbf{0.1735} & \textbf{2.9964} \\

 & LoRA & 47.84 & 0.3083 & 0.3057 & 0.3542 & 11.9778 & 37.36 & 0.2725 & 0.2700 & 0.4326 & 14.6916 \\

 & MixLoRA & \underline{49.06} & 0.2992 & 0.2790 & 0.3255 & 10.8608 & 36.74 & 0.2789 & 0.2593 & 0.4217 & 14.2837 \\

 & CuMA & \textbf{49.97} & \underline{0.2915} & \underline{0.2054} & \underline{0.2037} & 5.6714 & \textbf{38.58} & \textbf{0.2553} & 0.2064 & 0.2854 & 7.9169 \\

\rowcolor{paradeyellow}
\cellcolor{white}
 & \textbf{PARADE} & 48.34 & \textbf{0.2858} & \textbf{0.1270} & \textbf{0.1153} & \textbf{2.3284} & \underline{37.39} & \underline{0.2636} & \underline{0.1733} & \underline{0.1988} & \underline{3.8204} \\

\bottomrule
\end{tabular}%
}
\caption{Results for test-respondent resampling seed 46 on the standard WVS and SocioBench test splits. Bold and underlined values denote the best and second-best results within each backbone and dataset, respectively. The \method{} rows are highlighted in yellow.}
\label{tab:standard_seed_46_results}
\end{table*}

\subsection{Unseen Demographic Combinations}

Tables~\ref{tab:unseen_seed_42_results}--\ref{tab:unseen_seed_46_results}
give the corresponding per-seed evaluations on the unseen-combination split.
No method is
retrained for these evaluations. Every target eight-way tuple is absent from
training, while all eight constituent attribute values are observed; the
split therefore specifically evaluates compositional generalization rather
than extrapolation to new attribute values.

\begin{table*}[t]
\small
\centering
\setlength{\tabcolsep}{3.2pt}
\resizebox{\textwidth}{!}{%
\begin{tabular}{@{}llcccccccccc@{}}
\toprule
\textbf{Backbone}
& \textbf{Method}
& \multicolumn{5}{c}{\textbf{WVS}}
& \multicolumn{5}{c}{\textbf{SocioBench}} \\
\cmidrule(lr){3-7}\cmidrule(lr){8-12}
&
& \textbf{Acc. (\%) $\uparrow$}
& \textbf{MAE $\downarrow$}
& \textbf{EMD $\downarrow$}
& \textbf{JS $\downarrow$}
& \textbf{KL $\downarrow$}
& \textbf{Acc. (\%) $\uparrow$}
& \textbf{MAE $\downarrow$}
& \textbf{EMD $\downarrow$}
& \textbf{JS $\downarrow$}
& \textbf{KL $\downarrow$} \\
\midrule
\multirow{7}{*}{\shortstack{Llama-3.1-8B-\\Instruct}} & Vanilla Baseline & 37.94 & 0.3307 & 0.3305 & 0.4554 & 14.7521 & 33.51 & 0.2798 & 0.2723 & 0.4935 & 15.9756 \\

 & Persona Prompting & 36.75 & 0.3246 & 0.2727 & 0.3713 & 10.5249 & 35.79 & \underline{0.2565} & \underline{0.2165} & \underline{0.3582} & 10.9054 \\

 & SimVBG & 41.86 & 0.3345 & 0.2779 & 0.2835 & 7.3731 & 32.02 & 0.3033 & 0.2521 & 0.3665 & \underline{10.1624} \\

 & LoRA & \underline{46.67} & 0.3139 & 0.3079 & 0.3609 & 12.3256 & 36.93 & 0.2667 & 0.2587 & 0.4513 & 14.9653 \\

 & MixLoRA & \textbf{47.04} & \underline{0.3084} & 0.2866 & 0.3169 & 10.5383 & \underline{38.77} & 0.2571 & 0.2452 & 0.4223 & 14.1437 \\

 & CuMA & 45.18 & 0.3227 & \underline{0.2550} & \underline{0.2544} & \underline{7.3401} & 38.25 & 0.2602 & 0.2266 & 0.3773 & 11.7310 \\

\rowcolor{paradeyellow}
\cellcolor{white}
 & \textbf{PARADE} & 46.53 & \textbf{0.2969} & \textbf{0.1740} & \textbf{0.1884} & \textbf{5.3151} & \textbf{39.91} & \textbf{0.2563} & \textbf{0.1884} & \textbf{0.2794} & \textbf{8.5004} \\

\midrule

\multirow{7}{*}{Qwen3.5-9B} & Vanilla Baseline & 39.88 & 0.3501 & 0.3499 & 0.4481 & 14.0674 & 33.60 & 0.2989 & 0.2912 & 0.4906 & 15.9420 \\

 & Persona Prompting & 38.83 & 0.3360 & 0.2848 & 0.3500 & 8.5604 & 34.39 & 0.2721 & 0.2345 & 0.4106 & 12.6500 \\

 & SimVBG & 42.03 & 0.3163 & \underline{0.2223} & \underline{0.2267} & \underline{6.0686} & \underline{39.47} & \underline{0.2505} & \underline{0.1946} & \underline{0.2967} & \underline{8.8067} \\

 & LoRA & 45.54 & 0.3303 & 0.3290 & 0.3752 & 12.6149 & 37.63 & 0.2639 & 0.2483 & 0.4408 & 14.7154 \\

 & MixLoRA & 46.23 & 0.3232 & 0.3137 & 0.3564 & 11.9519 & 37.19 & 0.2718 & 0.2524 & 0.4303 & 14.3013 \\

 & CuMA & \underline{46.78} & \underline{0.3105} & 0.2450 & 0.2542 & 6.7641 & 38.68 & 0.2567 & 0.2173 & 0.3498 & 10.7860 \\

\rowcolor{paradeyellow}
\cellcolor{white}
 & \textbf{PARADE} & \textbf{47.24} & \textbf{0.3052} & \textbf{0.1881} & \textbf{0.2079} & \textbf{5.3956} & \textbf{39.91} & \textbf{0.2461} & \textbf{0.1767} & \textbf{0.2584} & \textbf{7.6043} \\

\midrule

\multirow{7}{*}{\shortstack{Ministral-3-8B-\\Instruct}} & Vanilla Baseline & 36.35 & 0.3351 & 0.3349 & 0.4852 & 15.1937 & 32.37 & 0.2839 & 0.2696 & 0.5049 & 16.2011 \\

 & Persona Prompting & 38.39 & 0.3087 & 0.2405 & 0.3055 & 7.6240 & 34.12 & 0.2621 & 0.2215 & 0.3818 & 11.3541 \\

 & SimVBG & 37.51 & 0.3312 & 0.2262 & 0.2263 & \underline{4.6027} & 35.44 & 0.2704 & \underline{0.1915} & \underline{0.2753} & \textbf{7.0827} \\

 & LoRA & 46.65 & 0.3201 & 0.3197 & 0.3683 & 12.4824 & 38.16 & 0.2629 & 0.2607 & 0.4416 & 14.7636 \\

 & MixLoRA & \underline{47.99} & 0.3099 & 0.2876 & 0.3334 & 11.2530 & 35.35 & 0.2809 & 0.2618 & 0.4551 & 14.7690 \\

 & CuMA & \textbf{48.66} & \textbf{0.3040} & \underline{0.2182} & \underline{0.2126} & 6.0327 & \textbf{39.39} & \textbf{0.2424} & 0.2058 & 0.3409 & 10.5771 \\

\rowcolor{paradeyellow}
\cellcolor{white}
 & \textbf{PARADE} & 46.61 & \underline{0.3047} & \textbf{0.1327} & \textbf{0.1181} & \textbf{2.5356} & \underline{38.95} & \underline{0.2554} & \textbf{0.1869} & \textbf{0.2698} & \underline{7.7406} \\

\bottomrule
\end{tabular}%
}
\caption{Results for test-respondent resampling seed 42 on the unseen-combination WVS and SocioBench splits. Bold and underlined values denote the best and second-best results within each backbone and dataset, respectively. The \method{} rows are highlighted in yellow.}
\label{tab:unseen_seed_42_results}
\end{table*}

\begin{table*}[t]
\small
\centering
\setlength{\tabcolsep}{3.2pt}
\resizebox{\textwidth}{!}{%
\begin{tabular}{@{}llcccccccccc@{}}
\toprule
\textbf{Backbone}
& \textbf{Method}
& \multicolumn{5}{c}{\textbf{WVS}}
& \multicolumn{5}{c}{\textbf{SocioBench}} \\
\cmidrule(lr){3-7}\cmidrule(lr){8-12}
&
& \textbf{Acc. (\%) $\uparrow$}
& \textbf{MAE $\downarrow$}
& \textbf{EMD $\downarrow$}
& \textbf{JS $\downarrow$}
& \textbf{KL $\downarrow$}
& \textbf{Acc. (\%) $\uparrow$}
& \textbf{MAE $\downarrow$}
& \textbf{EMD $\downarrow$}
& \textbf{JS $\downarrow$}
& \textbf{KL $\downarrow$} \\
\midrule
\multirow{7}{*}{\shortstack{Llama-3.1-8B-\\Instruct}} & Vanilla Baseline & 37.69 & 0.3295 & 0.3278 & 0.4590 & 14.7661 & 31.93 & 0.2925 & 0.2861 & 0.5127 & 16.3967 \\

 & Persona Prompting & 37.50 & 0.3161 & 0.2679 & 0.3621 & 10.0993 & 34.65 & \underline{0.2607} & 0.2236 & 0.3780 & 11.3502 \\

 & SimVBG & 42.00 & 0.3322 & 0.2697 & 0.2840 & 7.3560 & 29.21 & 0.3174 & 0.2664 & 0.3920 & \underline{10.7573} \\

 & LoRA & 46.19 & 0.3146 & 0.3091 & 0.3685 & 12.4572 & 35.88 & 0.2724 & 0.2662 & 0.4602 & 15.3086 \\

 & MixLoRA & \underline{46.84} & \underline{0.3091} & 0.2845 & 0.3178 & 10.4770 & 36.05 & 0.2682 & 0.2554 & 0.4387 & 14.6752 \\

 & CuMA & 46.09 & 0.3143 & \underline{0.2566} & \underline{0.2485} & \underline{6.9357} & \textbf{38.25} & \textbf{0.2597} & \underline{0.2218} & \underline{0.3586} & 11.7757 \\

\rowcolor{paradeyellow}
\cellcolor{white}
 & \textbf{PARADE} & \textbf{47.77} & \textbf{0.2879} & \textbf{0.1700} & \textbf{0.1831} & \textbf{4.8421} & \underline{36.93} & 0.2654 & \textbf{0.2032} & \textbf{0.3163} & \textbf{10.0339} \\

\midrule

\multirow{7}{*}{Qwen3.5-9B} & Vanilla Baseline & 39.78 & 0.3459 & 0.3442 & 0.4474 & 14.2041 & 29.47 & 0.3094 & 0.3035 & 0.5363 & 17.0657 \\

 & Persona Prompting & 39.48 & 0.3285 & 0.2760 & 0.3402 & 8.3004 & 32.81 & 0.2756 & 0.2328 & 0.4039 & 12.3784 \\

 & SimVBG & 43.28 & 0.3059 & \underline{0.2217} & \underline{0.2223} & \underline{5.7614} & 35.70 & \underline{0.2614} & \underline{0.2041} & \underline{0.3142} & \underline{9.0076} \\

 & LoRA & 46.04 & 0.3232 & 0.3214 & 0.3709 & 12.4508 & \underline{37.46} & 0.2675 & 0.2546 & 0.4429 & 14.7376 \\

 & MixLoRA & 46.35 & 0.3173 & 0.3022 & 0.3530 & 11.8209 & 36.32 & 0.2811 & 0.2645 & 0.4447 & 14.7111 \\

 & CuMA & \underline{47.92} & \underline{0.2962} & 0.2353 & 0.2486 & 6.9602 & 37.19 & 0.2624 & 0.2277 & 0.3667 & 11.6222 \\

\rowcolor{paradeyellow}
\cellcolor{white}
 & \textbf{PARADE} & \textbf{48.51} & \textbf{0.2875} & \textbf{0.1713} & \textbf{0.2008} & \textbf{5.4338} & \textbf{38.51} & \textbf{0.2490} & \textbf{0.1758} & \textbf{0.2603} & \textbf{7.5472} \\

\midrule

\multirow{7}{*}{\shortstack{Ministral-3-8B-\\Instruct}} & Vanilla Baseline & 36.22 & 0.3320 & 0.3306 & 0.4836 & 15.1831 & 30.88 & 0.2856 & 0.2721 & 0.5112 & 16.5939 \\

 & Persona Prompting & 38.89 & \underline{0.3018} & 0.2337 & 0.2963 & 7.4168 & 31.67 & \underline{0.2666} & 0.2153 & 0.3963 & 11.9529 \\

 & SimVBG & 38.57 & 0.3174 & 0.2315 & 0.2355 & \underline{4.8616} & 32.46 & 0.2759 & \textbf{0.1902} & \textbf{0.2826} & \textbf{7.6979} \\

 & LoRA & 46.61 & 0.3137 & 0.3130 & 0.3696 & 12.4698 & 34.82 & 0.2761 & 0.2740 & 0.4716 & 15.6346 \\

 & MixLoRA & \underline{48.04} & 0.3042 & 0.2823 & 0.3332 & 11.1716 & 34.82 & 0.2857 & 0.2681 & 0.4535 & 15.0109 \\

 & CuMA & \textbf{49.03} & 0.3030 & \underline{0.2170} & \underline{0.2134} & 6.0399 & \textbf{37.89} & \textbf{0.2518} & 0.2158 & 0.3345 & 10.0591 \\

\rowcolor{paradeyellow}
\cellcolor{white}
 & \textbf{PARADE} & 47.41 & \textbf{0.2983} & \textbf{0.1386} & \textbf{0.1214} & \textbf{2.2767} & \underline{35.96} & 0.2698 & \underline{0.2085} & \underline{0.3074} & \underline{9.3360} \\

\bottomrule
\end{tabular}%
}
\caption{Results for test-respondent resampling seed 43 on the unseen-combination WVS and SocioBench splits. Bold and underlined values denote the best and second-best results within each backbone and dataset, respectively. The \method{} rows are highlighted in yellow.}
\label{tab:unseen_seed_43_results}
\end{table*}

\begin{table*}[t]
\small
\centering
\setlength{\tabcolsep}{3.2pt}
\resizebox{\textwidth}{!}{%
\begin{tabular}{@{}llcccccccccc@{}}
\toprule
\textbf{Backbone}
& \textbf{Method}
& \multicolumn{5}{c}{\textbf{WVS}}
& \multicolumn{5}{c}{\textbf{SocioBench}} \\
\cmidrule(lr){3-7}\cmidrule(lr){8-12}
&
& \textbf{Acc. (\%) $\uparrow$}
& \textbf{MAE $\downarrow$}
& \textbf{EMD $\downarrow$}
& \textbf{JS $\downarrow$}
& \textbf{KL $\downarrow$}
& \textbf{Acc. (\%) $\uparrow$}
& \textbf{MAE $\downarrow$}
& \textbf{EMD $\downarrow$}
& \textbf{JS $\downarrow$}
& \textbf{KL $\downarrow$} \\
\midrule
\multirow{7}{*}{\shortstack{Llama-3.1-8B-\\Instruct}} & Vanilla Baseline & 37.90 & 0.3255 & 0.3251 & 0.4593 & 14.7956 & 31.58 & 0.2995 & 0.2928 & 0.5142 & 16.4754 \\

 & Persona Prompting & 37.71 & 0.3114 & 0.2624 & 0.3615 & 9.6384 & 35.79 & \underline{0.2663} & \underline{0.2274} & 0.3540 & 10.7100 \\

 & SimVBG & 43.18 & 0.3230 & 0.2643 & 0.2753 & \underline{7.0596} & 32.02 & 0.3142 & 0.2502 & \underline{0.3527} & \underline{9.9100} \\

 & LoRA & 47.73 & 0.3041 & 0.2992 & 0.3533 & 12.1021 & 36.32 & 0.2802 & 0.2742 & 0.4594 & 15.1695 \\

 & MixLoRA & \underline{47.81} & \underline{0.2992} & 0.2767 & 0.3086 & 10.2214 & 35.96 & 0.2759 & 0.2638 & 0.4430 & 14.6857 \\

 & CuMA & 46.18 & 0.3118 & \underline{0.2528} & \underline{0.2479} & 7.2152 & \textbf{38.51} & \textbf{0.2657} & 0.2295 & 0.3616 & 11.6895 \\

\rowcolor{paradeyellow}
\cellcolor{white}
 & \textbf{PARADE} & \textbf{48.20} & \textbf{0.2797} & \textbf{0.1679} & \textbf{0.1768} & \textbf{4.9605} & \underline{37.19} & 0.2726 & \textbf{0.2070} & \textbf{0.3057} & \textbf{9.8609} \\

\midrule

\multirow{7}{*}{Qwen3.5-9B} & Vanilla Baseline & 40.56 & 0.3397 & 0.3391 & 0.4424 & 14.1236 & 30.70 & 0.3084 & 0.3027 & 0.5245 & 16.7341 \\

 & Persona Prompting & 40.05 & 0.3209 & 0.2704 & 0.3387 & 8.2468 & 32.46 & 0.2708 & 0.2349 & 0.4199 & 13.0378 \\

 & SimVBG & 43.78 & 0.2979 & \underline{0.2153} & \underline{0.2187} & \underline{5.4953} & 35.61 & \textbf{0.2577} & \underline{0.2031} & \underline{0.3157} & \underline{9.0738} \\

 & LoRA & 46.38 & 0.3227 & 0.3216 & 0.3690 & 12.4284 & 37.28 & 0.2682 & 0.2546 & 0.4452 & 14.8185 \\

 & MixLoRA & 47.43 & 0.3127 & 0.3007 & 0.3473 & 11.6932 & 35.61 & 0.2866 & 0.2683 & 0.4465 & 14.6615 \\

 & CuMA & \underline{48.20} & \underline{0.2940} & 0.2299 & 0.2449 & 6.7631 & \underline{37.54} & 0.2651 & 0.2318 & 0.3688 & 11.7604 \\

\rowcolor{paradeyellow}
\cellcolor{white}
 & \textbf{PARADE} & \textbf{48.68} & \textbf{0.2836} & \textbf{0.1654} & \textbf{0.1932} & \textbf{5.2365} & \textbf{38.51} & \underline{0.2612} & \textbf{0.1881} & \textbf{0.2780} & \textbf{8.4145} \\

\midrule

\multirow{7}{*}{\shortstack{Ministral-3-8B-\\Instruct}} & Vanilla Baseline & 36.99 & 0.3267 & 0.3265 & 0.4772 & 15.0619 & 30.96 & 0.2935 & 0.2795 & 0.5161 & 16.5441 \\

 & Persona Prompting & 39.17 & 0.2970 & 0.2342 & 0.2919 & 7.2186 & 33.33 & 0.2664 & 0.2137 & 0.3768 & 11.9185 \\

 & SimVBG & 38.24 & 0.3138 & 0.2221 & 0.2264 & \underline{4.4469} & 35.96 & \underline{0.2608} & \textbf{0.1882} & \textbf{0.2726} & \textbf{7.5035} \\

 & LoRA & 47.37 & 0.3111 & 0.3111 & 0.3638 & 12.3368 & 35.70 & 0.2856 & 0.2823 & 0.4622 & 15.3269 \\

 & MixLoRA & \underline{48.46} & 0.3038 & 0.2828 & 0.3286 & 11.0497 & 34.65 & 0.2926 & 0.2746 & 0.4594 & 15.0707 \\

 & CuMA & \textbf{49.55} & \textbf{0.2957} & \underline{0.2122} & \underline{0.2056} & 5.5261 & \textbf{38.77} & \textbf{0.2566} & 0.2235 & 0.3341 & 10.4196 \\

\rowcolor{paradeyellow}
\cellcolor{white}
 & \textbf{PARADE} & 47.47 & \underline{0.2962} & \textbf{0.1355} & \textbf{0.1164} & \textbf{2.2913} & \underline{36.32} & 0.2770 & \underline{0.2083} & \underline{0.2952} & \underline{8.7815} \\

\bottomrule
\end{tabular}%
}
\caption{Results for test-respondent resampling seed 44 on the unseen-combination WVS and SocioBench splits. Bold and underlined values denote the best and second-best results within each backbone and dataset, respectively. The \method{} rows are highlighted in yellow.}
\label{tab:unseen_seed_44_results}
\end{table*}

\begin{table*}[t]
\small
\centering
\setlength{\tabcolsep}{3.2pt}
\resizebox{\textwidth}{!}{%
\begin{tabular}{@{}llcccccccccc@{}}
\toprule
\textbf{Backbone}
& \textbf{Method}
& \multicolumn{5}{c}{\textbf{WVS}}
& \multicolumn{5}{c}{\textbf{SocioBench}} \\
\cmidrule(lr){3-7}\cmidrule(lr){8-12}
&
& \textbf{Acc. (\%) $\uparrow$}
& \textbf{MAE $\downarrow$}
& \textbf{EMD $\downarrow$}
& \textbf{JS $\downarrow$}
& \textbf{KL $\downarrow$}
& \textbf{Acc. (\%) $\uparrow$}
& \textbf{MAE $\downarrow$}
& \textbf{EMD $\downarrow$}
& \textbf{JS $\downarrow$}
& \textbf{KL $\downarrow$} \\
\midrule
\multirow{7}{*}{\shortstack{Llama-3.1-8B-\\Instruct}} & Vanilla Baseline & 37.55 & 0.3282 & 0.3274 & 0.4608 & 14.8343 & 31.58 & 0.3029 & 0.2961 & 0.5270 & 16.4861 \\

 & Persona Prompting & 37.88 & 0.3112 & 0.2602 & 0.3604 & 10.4011 & 33.60 & 0.2773 & \underline{0.2382} & \underline{0.3824} & 11.0448 \\

 & SimVBG & 42.51 & 0.3272 & 0.2670 & 0.2789 & 7.4148 & 30.09 & 0.3294 & 0.2716 & 0.3982 & \underline{10.8819} \\

 & LoRA & 48.05 & 0.3057 & 0.2996 & 0.3499 & 11.9463 & 37.72 & 0.2693 & 0.2633 & 0.4508 & 14.7847 \\

 & MixLoRA & \underline{48.07} & \underline{0.3041} & 0.2818 & 0.3096 & 10.1983 & \textbf{39.04} & \textbf{0.2662} & 0.2561 & 0.4281 & 14.1539 \\

 & CuMA & 46.99 & 0.3062 & \underline{0.2440} & \underline{0.2418} & \underline{7.1768} & 37.28 & 0.2725 & 0.2399 & 0.3828 & 11.9445 \\

\rowcolor{paradeyellow}
\cellcolor{white}
 & \textbf{PARADE} & \textbf{48.27} & \textbf{0.2853} & \textbf{0.1689} & \textbf{0.1805} & \textbf{5.1062} & \underline{38.42} & \underline{0.2692} & \textbf{0.2014} & \textbf{0.2994} & \textbf{9.0957} \\

\midrule

\multirow{7}{*}{Qwen3.5-9B} & Vanilla Baseline & 40.08 & 0.3404 & 0.3397 & 0.4466 & 14.1055 & 32.72 & 0.2943 & 0.2886 & 0.5094 & 16.2091 \\

 & Persona Prompting & 40.04 & 0.3193 & 0.2686 & 0.3349 & 8.0780 & 34.39 & 0.2748 & 0.2319 & 0.4046 & 12.2464 \\

 & SimVBG & 43.18 & 0.3016 & \underline{0.2156} & \underline{0.2215} & \underline{5.7071} & 34.91 & 0.2782 & \underline{0.2091} & \underline{0.3100} & \underline{8.5014} \\

 & LoRA & 46.54 & 0.3247 & 0.3229 & 0.3679 & 12.3288 & \underline{38.25} & \underline{0.2659} & 0.2547 & 0.4418 & 14.6322 \\

 & MixLoRA & 47.60 & 0.3151 & 0.3036 & 0.3465 & 11.5784 & 37.46 & 0.2778 & 0.2617 & 0.4411 & 14.4882 \\

 & CuMA & \underline{47.80} & \underline{0.2992} & 0.2316 & 0.2472 & 6.3237 & 38.07 & 0.2727 & 0.2367 & 0.3652 & 11.0831 \\

\rowcolor{paradeyellow}
\cellcolor{white}
 & \textbf{PARADE} & \textbf{49.07} & \textbf{0.2882} & \textbf{0.1690} & \textbf{0.1924} & \textbf{5.3623} & \textbf{41.14} & \textbf{0.2481} & \textbf{0.1928} & \textbf{0.2680} & \textbf{7.6610} \\

\midrule

\multirow{7}{*}{\shortstack{Ministral-3-8B-\\Instruct}} & Vanilla Baseline & 36.51 & 0.3309 & 0.3302 & 0.4823 & 15.1372 & 32.54 & 0.2794 & 0.2657 & 0.5098 & 16.1415 \\

 & Persona Prompting & 39.84 & \textbf{0.2940} & 0.2351 & 0.2957 & 7.3271 & 34.04 & \underline{0.2690} & 0.2255 & 0.3893 & 12.0296 \\

 & SimVBG & 38.41 & 0.3182 & 0.2213 & 0.2292 & \underline{4.8931} & 32.02 & 0.2971 & \textbf{0.1925} & \textbf{0.2673} & \textbf{6.8554} \\

 & LoRA & 47.19 & 0.3143 & 0.3140 & 0.3664 & 12.3534 & \textbf{38.77} & 0.2718 & 0.2688 & 0.4382 & 14.5749 \\

 & MixLoRA & \underline{48.13} & 0.3090 & 0.2854 & 0.3305 & 11.0586 & 37.89 & 0.2759 & 0.2629 & 0.4353 & 14.3187 \\

 & CuMA & \textbf{49.64} & \underline{0.2959} & \underline{0.2097} & \underline{0.2045} & 5.8480 & \underline{38.25} & \textbf{0.2672} & 0.2207 & 0.3446 & 10.6113 \\

\rowcolor{paradeyellow}
\cellcolor{white}
 & \textbf{PARADE} & 47.76 & 0.2962 & \textbf{0.1250} & \textbf{0.1190} & \textbf{2.5963} & 36.14 & 0.2803 & \underline{0.2110} & \underline{0.3120} & \underline{8.6753} \\

\bottomrule
\end{tabular}%
}
\caption{Results for test-respondent resampling seed 45 on the unseen-combination WVS and SocioBench splits. Bold and underlined values denote the best and second-best results within each backbone and dataset, respectively. The \method{} rows are highlighted in yellow.}
\label{tab:unseen_seed_45_results}
\end{table*}

\begin{table*}[t]
\small
\centering
\setlength{\tabcolsep}{3.2pt}
\resizebox{\textwidth}{!}{%
\begin{tabular}{@{}llcccccccccc@{}}
\toprule
\textbf{Backbone}
& \textbf{Method}
& \multicolumn{5}{c}{\textbf{WVS}}
& \multicolumn{5}{c}{\textbf{SocioBench}} \\
\cmidrule(lr){3-7}\cmidrule(lr){8-12}
&
& \textbf{Acc. (\%) $\uparrow$}
& \textbf{MAE $\downarrow$}
& \textbf{EMD $\downarrow$}
& \textbf{JS $\downarrow$}
& \textbf{KL $\downarrow$}
& \textbf{Acc. (\%) $\uparrow$}
& \textbf{MAE $\downarrow$}
& \textbf{EMD $\downarrow$}
& \textbf{JS $\downarrow$}
& \textbf{KL $\downarrow$} \\
\midrule
\multirow{7}{*}{\shortstack{Llama-3.1-8B-\\Instruct}} & Vanilla Baseline & 37.72 & 0.3258 & 0.3252 & 0.4602 & 14.7551 & 31.49 & 0.2997 & 0.2933 & 0.5197 & 16.5514 \\

 & Persona Prompting & 37.66 & 0.3109 & 0.2606 & 0.3555 & 9.9201 & 34.21 & 0.2687 & \underline{0.2203} & 0.3685 & 11.0880 \\

 & SimVBG & 42.29 & 0.3247 & 0.2644 & 0.2733 & \underline{7.0454} & 30.88 & 0.3186 & 0.2476 & 0.3627 & \underline{9.7298} \\

 & LoRA & \underline{47.08} & 0.3043 & 0.2978 & 0.3546 & 11.8258 & 37.54 & \underline{0.2618} & 0.2555 & 0.4481 & 14.8569 \\

 & MixLoRA & \textbf{47.55} & \underline{0.2997} & 0.2775 & 0.3114 & 10.3122 & 37.28 & 0.2633 & 0.2552 & 0.4460 & 14.6137 \\

 & CuMA & 46.90 & 0.3055 & \underline{0.2462} & \underline{0.2476} & 7.1231 & \underline{37.81} & 0.2670 & 0.2207 & \underline{0.3574} & 11.2270 \\

\rowcolor{paradeyellow}
\cellcolor{white}
 & \textbf{PARADE} & \underline{47.08} & \textbf{0.2876} & \textbf{0.1654} & \textbf{0.1809} & \textbf{4.7545} & \textbf{38.77} & \textbf{0.2561} & \textbf{0.1989} & \textbf{0.3058} & \textbf{9.3288} \\

\midrule

\multirow{7}{*}{Qwen3.5-9B} & Vanilla Baseline & 38.86 & 0.3462 & 0.3455 & 0.4599 & 14.2543 & 31.40 & 0.3004 & 0.2945 & 0.5270 & 16.5866 \\

 & Persona Prompting & 39.64 & 0.3239 & 0.2698 & 0.3341 & 8.1256 & 31.49 & 0.2771 & 0.2363 & 0.4370 & 12.9845 \\

 & SimVBG & 43.56 & 0.3038 & \underline{0.2099} & \underline{0.2047} & \textbf{5.3859} & 35.61 & 0.2625 & \underline{0.1966} & \underline{0.3115} & \underline{9.0171} \\

 & LoRA & 46.41 & 0.3187 & 0.3173 & 0.3652 & 12.2543 & \underline{38.16} & \underline{0.2591} & 0.2457 & 0.4431 & 14.6637 \\

 & MixLoRA & 46.78 & 0.3121 & 0.2979 & 0.3481 & 11.7229 & 36.84 & 0.2697 & 0.2496 & 0.4355 & 14.3686 \\

 & CuMA & \underline{47.63} & \underline{0.2977} & 0.2253 & 0.2412 & 6.6634 & 36.84 & 0.2637 & 0.2248 & 0.3605 & 11.2885 \\

\rowcolor{paradeyellow}
\cellcolor{white}
 & \textbf{PARADE} & \textbf{47.98} & \textbf{0.2888} & \textbf{0.1679} & \textbf{0.1982} & \underline{5.4523} & \textbf{38.25} & \textbf{0.2471} & \textbf{0.1825} & \textbf{0.2712} & \textbf{8.1692} \\

\midrule

\multirow{7}{*}{\shortstack{Ministral-3-8B-\\Instruct}} & Vanilla Baseline & 36.35 & 0.3329 & 0.3324 & 0.4868 & 15.1397 & 32.81 & 0.2771 & 0.2627 & 0.4964 & 16.0868 \\

 & Persona Prompting & 39.25 & 0.2990 & 0.2320 & 0.2918 & 7.4233 & 31.75 & 0.2686 & 0.2087 & 0.3838 & 11.7612 \\

 & SimVBG & 38.14 & 0.3194 & 0.2211 & 0.2183 & \underline{4.3467} & 33.33 & 0.2837 & \textbf{0.1924} & \textbf{0.2626} & \textbf{6.9812} \\

 & LoRA & 47.27 & 0.3091 & 0.3089 & 0.3610 & 12.1127 & \underline{36.93} & 0.2669 & 0.2635 & 0.4571 & 15.0913 \\

 & MixLoRA & \underline{48.47} & 0.2990 & 0.2799 & 0.3283 & 11.0735 & 36.58 & 0.2738 & 0.2557 & 0.4416 & 14.6166 \\

 & CuMA & \textbf{48.86} & \underline{0.2939} & \underline{0.2068} & \underline{0.2051} & 5.7538 & \underline{36.93} & \underline{0.2599} & 0.2134 & 0.3481 & 10.9034 \\

\rowcolor{paradeyellow}
\cellcolor{white}
 & \textbf{PARADE} & 48.10 & \textbf{0.2890} & \textbf{0.1233} & \textbf{0.1182} & \textbf{2.5387} & \textbf{38.25} & \textbf{0.2587} & \underline{0.1962} & \underline{0.3058} & \underline{9.0857} \\

\bottomrule
\end{tabular}%
}
\caption{Results for test-respondent resampling seed 46 on the unseen-combination WVS and SocioBench splits. Bold and underlined values denote the best and second-best results within each backbone and dataset, respectively. The \method{} rows are highlighted in yellow.}
\label{tab:unseen_seed_46_results}
\end{table*}

\subsection{Parameter Sensitivity}

Tables~\ref{tab:lambda_validation_wvs} and
\ref{tab:lambda_validation_sociobench} report validation performance across
orthogonality coefficients. The final coefficient and composition scales for
each dataset--backbone setting are selected on the validation split by jointly
considering Accuracy, MAE, EMD, JS, and KL. No test result is used for this
selection.

\begin{table*}[t]
\small
\centering
\setlength{\tabcolsep}{3.2pt}
\renewcommand{\arraystretch}{1.08}
\begin{tabular}{@{}C{0.18\textwidth}C{0.065\textwidth}C{0.115\textwidth}C{0.10\textwidth}C{0.10\textwidth}C{0.10\textwidth}C{0.10\textwidth}@{}}
\toprule
\textbf{Backbone} & $\boldsymbol{\lambda}$ & \textbf{Acc. (\%) $\uparrow$} & \textbf{MAE $\downarrow$} & \textbf{EMD $\downarrow$} & \textbf{JS $\downarrow$} & \textbf{KL $\downarrow$} \\
\midrule
\multirow{6}{0.18\textwidth}{\centering Llama-3.1-8B-Instruct}
& 0   & 46.92 & 0.2892 & 0.1796 & 0.1868 & 4.6744 \\
& 0.2 & 46.65 & 0.2952 & 0.1727 & 0.1742 & 4.2061 \\
& \selectedcell{\textbf{0.3}} & \selectedcell{47.06} & \selectedcell{0.2891} & \selectedcell{0.1728} & \selectedcell{0.1826} & \selectedcell{4.9778} \\
& 0.4 & 46.02 & 0.2974 & 0.1765 & 0.1843 & 4.8392 \\
& 0.5 & 46.72 & 0.2955 & 0.1640 & 0.1721 & 4.2094 \\
& 1.0 & 46.70 & 0.2909 & 0.1740 & 0.1851 & 4.5194 \\
\midrule
\multirow{6}{0.18\textwidth}{\centering Qwen3.5-9B}
& 0   & 48.17 & 0.2935 & 0.1790 & 0.1906 & 4.6466 \\
& 0.4 & 48.27 & 0.2915 & 0.1817 & 0.1931 & 5.0431 \\
& \selectedcell{\textbf{0.5}} & \selectedcell{48.76} & \selectedcell{0.2884} & \selectedcell{0.1765} & \selectedcell{0.1901} & \selectedcell{4.8473} \\
& 0.6 & 48.68 & 0.2885 & 0.1872 & 0.2019 & 5.2831 \\
& 0.7 & 47.70 & 0.2964 & 0.1771 & 0.1857 & 4.8531 \\
& 1.0 & 48.35 & 0.2898 & 0.1787 & 0.1896 & 4.8956 \\
\midrule
\multirow{6}{0.18\textwidth}{\centering Ministral-3-8B-Instruct}
& 0   & 47.82 & 0.2965 & 0.1749 & 0.1818 & 4.6468 \\
& 0.1 & 47.10 & 0.2991 & 0.1615 & 0.1653 & 3.7416 \\
& 0.2 & 47.76 & 0.2937 & 0.1613 & 0.1636 & 3.8839 \\
& \selectedcell{\textbf{0.3}} & \selectedcell{48.03} & \selectedcell{0.2915} & \selectedcell{0.1247} & \selectedcell{0.1177} & \selectedcell{2.3016} \\
& 0.5 & 47.19 & 0.2983 & 0.1836 & 0.1887 & 4.1185 \\
& 1.0 & 47.47 & 0.2987 & 0.1831 & 0.1841 & 4.2810 \\
\bottomrule
\end{tabular}
\caption{Validation-set sensitivity to the orthogonality coefficient on
WVS with unit Task and Demographic scales. Yellow cells denote the
coefficients used in the reported experiments after validation-based selection
jointly with the composition scales.}
\label{tab:lambda_validation_wvs}
\end{table*}

\begin{table*}[t]
\small
\centering
\setlength{\tabcolsep}{3.2pt}
\renewcommand{\arraystretch}{1.08}
\begin{tabular}{@{}C{0.18\textwidth}C{0.065\textwidth}C{0.115\textwidth}C{0.10\textwidth}C{0.10\textwidth}C{0.10\textwidth}C{0.10\textwidth}@{}}
\toprule
\textbf{Backbone} & $\boldsymbol{\lambda}$ & \textbf{Acc. (\%) $\uparrow$} & \textbf{MAE $\downarrow$} & \textbf{EMD $\downarrow$} & \textbf{JS $\downarrow$} & \textbf{KL $\downarrow$} \\
\midrule
\multirow{11}{0.18\textwidth}{\centering Llama-3.1-8B-Instruct}
& 0.0 & 37.04 & 0.2667 & 0.1856 & 0.2310 & 5.8511 \\
& 0.1 & 36.84 & 0.2664 & 0.1875 & 0.2387 & 6.3168 \\
& 0.2 & 37.35 & 0.2675 & 0.1811 & 0.2325 & 6.3164 \\
& 0.3 & 36.47 & 0.2689 & 0.1797 & 0.2241 & 5.9182 \\
& \selectedcell{\textbf{0.4}} & \selectedcell{36.70} & \selectedcell{0.2729} & \selectedcell{0.1908} & \selectedcell{0.2405} & \selectedcell{6.5069} \\
& 0.5 & 36.96 & 0.2668 & 0.1884 & 0.2365 & 6.1032 \\
& 0.6 & 36.93 & 0.2673 & 0.1890 & 0.2422 & 6.4635 \\
& 0.7 & 37.18 & 0.2642 & 0.1880 & 0.2395 & 6.6326 \\
& 0.8 & 36.32 & 0.2725 & 0.1899 & 0.2377 & 6.4433 \\
& 0.9 & 36.75 & 0.2688 & 0.1826 & 0.2286 & 6.0714 \\
& 1.0 & 37.16 & 0.2661 & 0.1891 & 0.2356 & 6.1553 \\
\midrule
\multirow{7}{0.18\textwidth}{\centering Qwen3.5-9B}
& 0   & 37.30 & 0.2658 & 0.1886 & 0.2351 & 6.2668 \\
& 0.5 & 37.21 & 0.2668 & 0.1855 & 0.2418 & 6.5583 \\
& 0.6 & 39.05 & 0.2521 & 0.1753 & 0.2228 & 5.8753 \\
& \selectedcell{\textbf{0.7}} & \selectedcell{38.88} & \selectedcell{0.2508} & \selectedcell{0.1695} & \selectedcell{0.2188} & \selectedcell{5.6481} \\
& 0.8 & 37.16 & 0.2659 & 0.1821 & 0.2339 & 6.3932 \\
& 0.9 & 38.79 & 0.2523 & 0.1752 & 0.2239 & 5.8355 \\
& 1.0 & 37.72 & 0.2643 & 0.1860 & 0.2404 & 6.4214 \\
\midrule
\multirow{11}{0.18\textwidth}{\centering Ministral-3-8B-Instruct}
& 0    & 36.98 & 0.2666 & 0.1968 & 0.2411 & 5.9226 \\
& 0.1  & 37.21 & 0.2652 & 0.1922 & 0.2389 & 6.0485 \\
& 0.2  & 37.35 & 0.2633 & 0.1862 & 0.2279 & 5.6438 \\
& 0.3  & 37.61 & 0.2619 & 0.1812 & 0.2197 & 5.4835 \\
& 0.4  & 37.28 & 0.2638 & 0.1862 & 0.2295 & 5.6239 \\
& \selectedcell{\textbf{0.5}} & \selectedcell{37.58} & \selectedcell{0.2599} & \selectedcell{0.1793} & \selectedcell{0.2205} & \selectedcell{5.3665} \\
& 0.6  & 37.53 & 0.2642 & 0.1901 & 0.2255 & 5.3923 \\
& 0.7  & 37.39 & 0.2626 & 0.1856 & 0.2234 & 5.2452 \\
& 0.8  & 36.91 & 0.2632 & 0.1837 & 0.2280 & 5.7362 \\
& 0.9  & 36.96 & 0.2632 & 0.1825 & 0.2150 & 4.9527 \\
& 1.0  & 37.37 & 0.2670 & 0.1856 & 0.2314 & 5.5322 \\
\bottomrule
\end{tabular}
\caption{Validation-set sensitivity to the orthogonality coefficient on
SocioBench with unit Task and Demographic scales. Yellow cells denote the
coefficients used in the reported experiments after validation-based selection
jointly with the composition scales.}
\label{tab:lambda_validation_sociobench}
\end{table*}

Performance is nonmonotonic in $\lambda$, and the preferred coefficient
varies across backbone and dataset, supporting validation-based selection
rather than a single coefficient shared by all settings.

\end{document}